\documentclass{article}

\PassOptionsToPackage{numbers, compress}{natbib}
\usepackage[final,eandd]{neurips_2026}

\usepackage[utf8]{inputenc} % allow utf-8 input
\usepackage[T1]{fontenc}    % use 8-bit T1 fonts
\ifdefined\pdfmapfile\pdfmapfile{+cm-super-t1.map}\fi
\usepackage[hidelinks]{hyperref}       % hyperlinks
\usepackage{url}            % simple URL typesetting
\usepackage{amsfonts}       % blackboard math symbols
\usepackage{nicefrac}       % compact symbols for 1/2, etc.
\usepackage{microtype}      % microtypography
\usepackage[table]{xcolor}         % colors
\usepackage{wrapfig}        % wrapfigure
\usepackage[nohyperlinks]{acronym} 
\usepackage{bbm}            % for binary indicator function
\usepackage{xspace}         % eg/etal onedot definition 
\usepackage{enumitem}       % bullet point in intro
\usepackage{bm}             % bold math symbols
\usepackage[skins, most]{tcolorbox}
\usepackage{multirow}
\usepackage{changepage} % for research question indent

\usepackage[textsize=tiny]{todonotes}

\usepackage{color}
\usepackage[small]{caption}
\usepackage{subcaption} % for subfigures

\usepackage{makecell} % newline in a table cell
\usepackage{fontawesome5}  % for warning icon
\usepackage{amssymb}
\usepackage{adjustbox}
\usepackage{amsthm}
\usepackage{amsmath}        % For theorems and declare math operator
\usepackage{listings}       % to render the code nicely
\usepackage{pifont}         % for ticking in tables

\usepackage{longtable}      % multipage task catalogue
\usepackage{booktabs}       % professional-quality tables
\usepackage{nicematrix}
\usepackage{arydshln}       % dashed table rules
\usepackage{bm}             % bold math

\usepackage{booktabs}
\makeatletter
\DeclareRobustCommand\onedot{\futurelet\@let@token\@onedot}
\def\@onedot{\ifx\@let@token.\else.\null\fi\xspace}

\def\eg{\emph{e.g}\onedot}

\makeatother

\newcommand{\ours}{\textsc{VeriWorld}\xspace}
\newcommand{\lean}{\textsc{Lean}\xspace}

\newcommand{\repourl}{\url{https://github.com/VITA-Group/VeriWorld}}

\newcommand{\cmark}{\ding{51}}
\newcommand{\xmark}{\ding{55}}

\definecolor{methodfg}{HTML}{FF8800}
\definecolor{methodbg}{RGB}{255,244,227}
\definecolor{lightgray}{HTML}{f0f0f0}
\definecolor{mypurple}{HTML}{A13FDB} % sft methods
\definecolor{mypink}{HTML}{E14B9E} % other methods
\definecolor{myblue}{HTML}{456AFF} % trigger words
\definecolor{myred}{HTML}{F20D0D} % unsafe
\definecolor{mygreen}{HTML}{1B8C2A} % dark green for tables
\definecolor{LightGray}{RGB}{235,235,235}
\definecolor{LightGreen}{RGB}{210,245,210}
\definecolor{LightRed}{RGB}{255,210,210}
\definecolor{LightBlue}{RGB}{180,205,245}
\definecolor{LightOrange}{RGB}{255,210,170}

\acrodef{lrm}[LRM]{large reasoning model}
\acrodef{cot}[CoT]{chain-of-thought}
\acrodef{sft}[SFT]{supervised finetuning}
\acrodef{rlhf}[RLHF]{reinforcement learning from human feedback}
\acrodef{dapo}[DAPO]{decouple clip and dynamic sampling policy optimization}
\acrodef{grpo}[GRPO]{group relative policy optimization }
\acrodef{rl}[RL]{reinforcement learning}
\acrodef{rlvr}[RLVR]{Reinforcement Learning with Verifiable Rewards}
\acrodef{ipr}[IPR]{iterative prefill reset}
\acrodef{mdp}[MDP]{markov decision process}
\acrodef{llm}[LLM]{large language model}
\acrodef{vlm}[VLM]{vision-language model}
\acrodef{ue}[UE]{Unreal Engine}

\title{\ours: A Verifiable Visual SWE-Bench for Spatial Reasoning in 3D Environments}
\author{%
\textbf{Yan Zheng\textsuperscript{1} \quad ShengYun Peng\textsuperscript{2} \quad Xuyao Liang\textsuperscript{3} \quad Xiaoyan Cong\textsuperscript{4}} \\
\textbf{Nuo Chen\textsuperscript{5} \quad Bangya Liu\textsuperscript{6} \quad Zihan Wang\textsuperscript{6} \quad Wenyan Cong\textsuperscript{1}} \\
\textbf{Zhiwen Fan\textsuperscript{5} \quad Zhangyang Wang\textsuperscript{1}} \\
\textnormal{\textsuperscript{1}The University of Texas at Austin} \\
\textnormal{\textsuperscript{2}Georgia Institute of Technology \quad \textsuperscript{3}HANABIT} \\
\textnormal{\textsuperscript{4}Brown University \quad \textsuperscript{5}Texas A\&M University} \\
\textnormal{\textsuperscript{6}Abaka AI}
}

\begin{document}
\maketitle

\begin{abstract}

\Acp{vlm} are increasingly used for spatial tasks such as navigation and manipulation, where they must infer geometry and object relationships from visual input in order to act correctly.
Existing benchmarks usually test perception and reasoning in isolation, making it hard to tell whether failures come from visual understanding, downstream reasoning, or the bridge between them.
We introduce \textbf{\ours}, a benchmark for spatial reasoning in interactive 3D environments with executable actions and deterministic verification.
For task families where all three channels are well-defined, we evaluate matched \textit{visual}, \textit{structured}, and \textit{combined} input conditions on the same instance, environment, and verifier; families without a clean visual/structured/combined split are reported separately.
This design supports controlled comparisons of perception, reasoning, and their interaction.
Using this setup, we uncover two diagnostic findings.
First, models often perform much better when task-relevant information, such as object identity, current state, or target location, is given explicitly than when it must be read from the rendered scene, revealing a recurring visual-to-structure gap.
Second, action-space and harness design can change outcomes even under the same task and information condition, showing that interaction protocol is itself a confounding variable.
\ours is open-sourced to support reproducible and fine-grained evaluation of spatial reasoning in \acp{vlm}.
\par\smallskip\noindent\textbf{Code:} \href{https://github.com/VITA-Group/VeriWorld}{\texttt{github.com/VITA-Group/VeriWorld}}\\
\textbf{Dataset:} \href{https://huggingface.co/datasets/yanyanzheng96/VeriWorld}{\texttt{huggingface.co/datasets/yanyanzheng96/VeriWorld}}

\end{abstract}

\section{Introduction}
\label{sec:intro}

\begin{figure*}[!t]
  \centering
  \includegraphics[width=\textwidth]{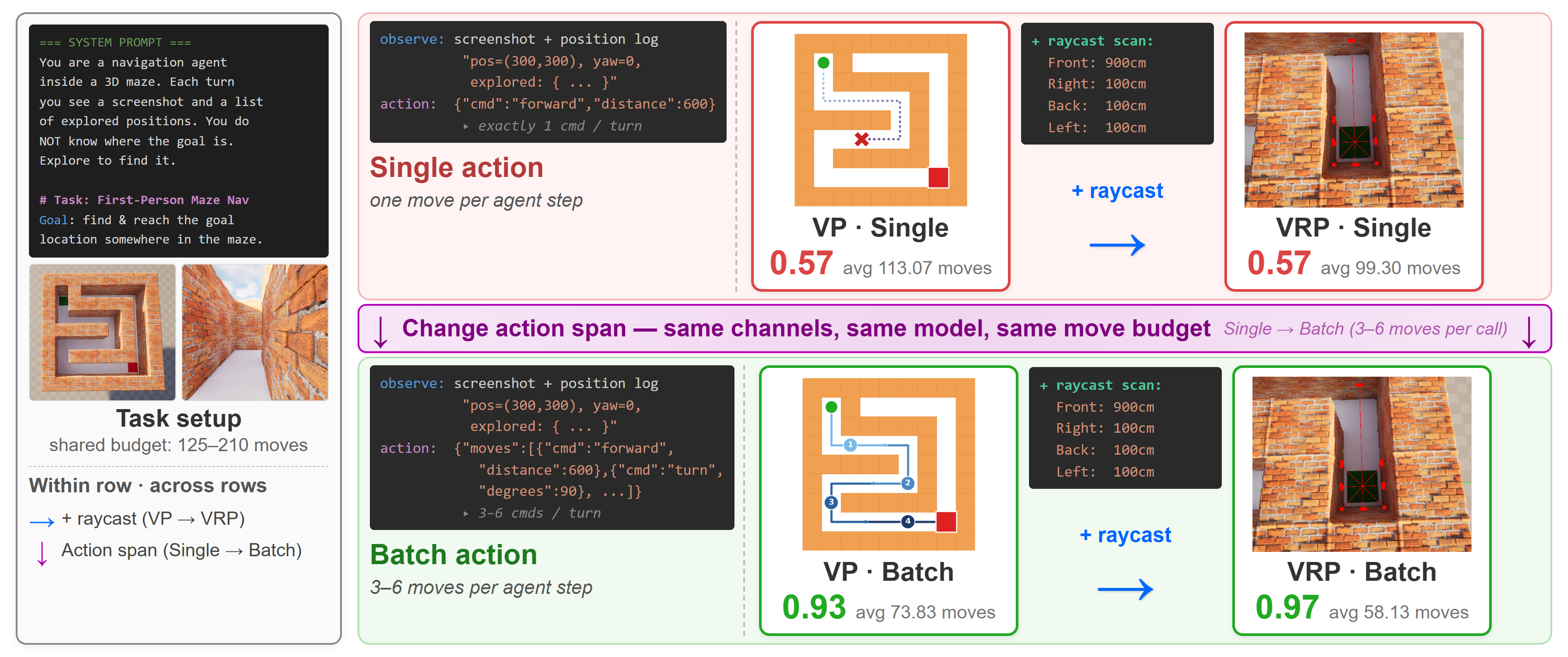}
  \caption{\textbf{Batch control improves performance under both evaluated information channels at an equal move budget.}
  Same task (MazeNavFPS, 30 mazes), same model (Gemini-3.1-Pro), and the same budget of primitive moves for both protocols (125/170/210 moves for $3{\times}3$/$4{\times}4$/$5{\times}5$ grids):
  switching the action interface from Single (one move per call) to Batch (3--6 moves per call) lifts Pass@1 from 0.57 to 0.93 under \textit{VP} (vision + position log) and from 0.57 to 0.97 under \textit{VRP} (\textit{VP} + raycast), while Batch uses fewer moves on average.
  Under an equal budget of 50 model calls, Batch also outperforms Single for every evaluated model (Sec.~\ref{sec:case_studies} and Table~\ref{tab:maze_call_budget}).
  This motivates \ours's three-axis design --- task, information channel, and action harness --- in which the harness is treated as a first-class experimental variable rather than a fixed implementation detail.}
  \label{fig:teaser_diagnose}
\end{figure*}

Vision-language models (\Acp{vlm})~\citep{zhao2024deep,tong2024cambrian,li2024llava,li2024mvbench,lin2024video,xu2025llava} are increasingly applied to reasoning about space, \eg, navigation, manipulation, and physical prediction.
In these settings, a model must recover task-relevant structure, such as object positions, distances, layout, connectivity, or physical and geometric parameters, from visual input, and use that structure not only to choose actions but also to refine them through environmental feedback.
For example, navigating a maze requires identifying free space and connectivity, while predicting a trajectory requires estimating geometry and distance.

However, existing benchmarks typically evaluate only isolated parts of this process.
Visual benchmarks such as SpatialVLM~\citep{chen2024spatialvlm}, GPT4Scene~\citep{qi2025gpt4scene}, and SIRI-Bench~\citep{song2025siri} focus on recognition or description from images, while reasoning benchmarks such as AlphaMaze~\citep{dao2025alphamaze} and Smooth Operator~\citep{jiao2026smooth} assume structured inputs such as numbers, equations, or symbolic expressions.
As a result, they do not cleanly distinguish whether failure comes from perception, downstream reasoning, or the interaction between them, even though real-world spatial tasks require all three to work together.
This leads to a central question:

\begin{tcolorbox}[colback=white, colframe=mypurple, width=0.99\textwidth, center, boxrule=1pt, arc=2pt]
\emph{What is the right way for a VLM to reason about space: through intuitive perceptual judgment, or through explicit symbolic reasoning with verifiable outcomes?}
\end{tcolorbox}

A useful benchmark for this question should satisfy three requirements.
First, it should be \emph{interactive}: the model should observe a scene, act, and receive feedback rather than answer a single static query.
Second, it should be \emph{verifiable}: success should be determined by a deterministic pass/fail criterion rather than subjective judgment.
Third, it should be \emph{diagnostic}: the same underlying task should reveal whether errors come from visual understanding, reasoning, or their interaction.

In this paper, we introduce \textbf{\ours}, a benchmark for evaluating spatial reasoning in \acp{vlm} through interactive tasks with verifiable outcomes.
Each task is instantiated as an interactive 3D environment, where a \ac{vlm} agent observes the scene, executes actions, and is evaluated by a binary pass/fail checker.
For task families where all three channels are well-defined, our core design runs the same task instance under three matched input conditions, visual, structured, and combined, while keeping the environment and objective fixed; tasks for which a clean V/S/C decomposition is not meaningful (photoreal scene grounding, pose-assisted navigation) are reported as specialized tracks in the appendix.
This controlled setup supports comparisons of how visual and structured inputs affect task performance.
Tasks are specified in a way that supports executable instantiation, deterministic checking, and parameterized generation (Fig.~\ref{fig:pipeline}). By comparing these configurations on the same task, \ours identifies cases for closer analysis of perception, reasoning, multimodal input use, and interaction protocol, beyond a single aggregate score.

 Our main contributions can be outlined as follows:

\begin{table}[t]
  \centering
\caption{
\textbf{\ours is the first benchmark to jointly support interactive 3D environments, executable code-as-action, deterministic verification, scalable parametric generation, and diagnostic evaluation through controlled input conditions.}
Prior benchmarks typically capture only subsets of these capabilities—focusing on perception, interaction, or evaluation in isolation—while \ours integrates all components within a unified and verifiable framework.
\cmark\ = supported, \xmark\ = not supported, $\sim$ = partial.
}
  \label{tab:comparison}
  \small
  \begin{tabular}{@{}l cccccc @{}}
    \toprule
    \textbf{Benchmark} & \textbf{Inter-} & \textbf{3D} & \textbf{Code-as-} & \textbf{Determ.} & \textbf{Param.} & \textbf{Diagnostic} \\
    & \textbf{active} & & \textbf{action} & \textbf{verif.} & \textbf{gen.} & \textbf{ablation} \\
    \midrule
    SpatialVLM~\citep{chen2024spatialvlm} & \xmark & \xmark & \xmark & \xmark & \cmark & \xmark \\
    GPT4Scene~\citep{qi2025gpt4scene} & \xmark & \cmark & \xmark & \xmark & \xmark & \xmark \\
    SIRI-Bench~\citep{song2025siri} & \xmark & \cmark & \xmark & $\sim$ & \cmark & \xmark \\
    lmgame-Bench~\citep{hu2025lmgame} & \cmark & \xmark & \xmark & \xmark & \xmark & \xmark \\
    ALFRED~\citep{shridhar2020alfred} & \cmark & \cmark & \xmark & $\sim$ & \xmark & \xmark \\
    Habitat~\citep{savva2019habitat} & \cmark & \cmark & \xmark & $\sim$ & $\sim$ & \xmark \\
    AlphaMaze~\citep{dao2025alphamaze} & \xmark & \xmark & \xmark & \cmark & \cmark & \xmark \\
    Smooth Operator~\citep{jiao2026smooth} & \xmark & \cmark & \xmark & $\sim$ & \cmark & \xmark \\
    VoxelCodeBench~\citep{zheng2026voxelcodebench} & \xmark & \cmark & \cmark & \xmark & \cmark & \xmark \\
    Sim2Reason~\citep{prabhudesai2026solving} & \xmark & \xmark & \xmark & \cmark & \cmark & \xmark \\
    GameWorld~\citep{ouyang2026gameworld} & \cmark & \xmark & \xmark & $\sim$ & \xmark & $\sim$ \\
    \midrule
    \textbf{\ours (ours)} & \cmark & \cmark & \cmark & \cmark & \cmark & \cmark \\
    \bottomrule
  \end{tabular}
\end{table}

\begin{enumerate}[leftmargin=*,topsep=0pt]
\itemsep0.2em
\item \textbf{We propose \ours, a benchmark for spatial reasoning in interactive 3D environments with deterministic, verifiable outcomes.}
\ours instantiates tasks as executable environments in which agents act and receive feedback, and evaluates them with a binary pass/fail checker.
Each task defines a goal condition, \eg, reaching a target state or satisfying geometric constraints, and the agent's final outcome is checked automatically against it to produce a clear binary signal.
This turns open-ended interaction into verifiable evaluation while supporting scalable experimentation through parameterized task generation and controlled variation of information channels and action interfaces.
We release \ours as a benchmark and diagnostic platform for reproducible spatial reasoning evaluation.

\item \textbf{We identify a recurring visual-to-structure gap through controlled input conditions on the same task.}
By evaluating the same tasks under visual, structured, and combined inputs, we show that models often succeed when task-relevant information is given explicitly but fail when it must be read from the rendered scene.
This identifies cases where explicit task information improves performance and motivates closer analysis of visual identification and state estimation.

\item \textbf{We show that interaction protocol is itself a confounding variable in spatial evaluation.}
Holding the task and information channel fixed, changing the action space or harness design can change the outcome from failure to success.
This demonstrates that benchmark conclusions can otherwise conflate information bottlenecks with action-space or protocol bottlenecks, and highlights the need to specify the inputs, instructions, and budgets behind each harness comparison.
\end{enumerate}
\section{Related Work}

\paragraph{Spatial reasoning and perception benchmarks.}
A large body of work evaluates spatial understanding in VLMs through recognition, question answering, or scene description on static images or videos. SpatialVLM~\citep{chen2024spatialvlm,cheng2024spatialrgpt,zhao2025cot,team2025gemini}, NLVR2~\citep{suhr2019nlvr2}, ScanQA~\citep{azuma2022scanqa}, and CLEVR~\citep{johnson2017clevr} probe compositional and relational reasoning; GPT4Scene~\citep{qi2025gpt4scene,wu2025spatial,zhu2025llava,yang2025mmsi}, VLM-3R~\citep{fan2025vlm3r}, Thinking-in-Space~\citep{yang2025thinking}, ScanRefer~\citep{chen2020scanrefer}, and SIRI-Bench~\citep{song2025siri} extend to 3D, video, or geometry-grounded settings; G2VLM~\citep{g2vlm2026} adds geometry-aware reasoning traces. Several already observe that performance improves when structured representations are exposed, but these benchmarks operate on static observations and natural-language outputs that are not tied to executable action and are weakly or non-deterministically verifiable. VoxelCodeBench~\citep{zheng2026voxelcodebench} moves toward executable 3D code generation but lacks interactive feedback loops and controlled attribution across information conditions. \ours instead evaluates spatial reasoning in an \emph{interactive, executable, and verifiable} setting where agents recover task-relevant structure through interaction and produce deterministically-checked actions --- shifting evaluation from passive perception to \emph{closed-loop reasoning and execution} (Table~\ref{tab:comparison}).

\paragraph{Embodied evaluation and verifiable environments.}
Embodied benchmarks evaluate agents acting in environments. ALFRED~\citep{shridhar2020alfred,li2024llavanext,ahn2022can}, Habitat~\citep{savva2019habitat,szot2021habitat}, AI2-THOR~\citep{kolve2017ai2thor}, and TEACh~\citep{padmakumar2022teach} target navigation and manipulation with task-completion metrics; lmgame-Bench~\citep{hu2025lmgame,lehrach2025code,xie2026memo} and MineDojo~\citep{fan2022minedojo} extend to open-ended games. These rely on success rates or heuristic rewards, making it hard to disentangle perception, reasoning, and integration failures. AlphaMaze~\citep{dao2025alphamaze} and related efforts add rule-based or programmatic checking. Two concurrent benchmarks bracket \ours: Sim2Reason~\citep{prabhudesai2026solving} synthesises verifiable physics-reasoning QA from MuJoCo but feeds the model only text-converted numerical traces --- the scene is never observed visually, so visual-spatial perception is untested. GameWorld~\citep{ouyang2026gameworld} compares two pre-defined action interfaces (Computer-Use vs.\ Semantic Action Parsing) across 34 browser games, but its pre-built games preclude holding task and information fixed while varying only the action protocol on a single task, and cannot introduce new action types such as code or shader programs. \ours controls task construction end-to-end, enabling comparisons of action protocols on the \emph{same} task instance under specified interaction budgets (Sec.~\ref{sec:case_studies}). Its visual, structured, and combined input conditions support task-specific analysis of perception, reasoning, multimodal input use, and interaction protocol (Table~\ref{tab:comparison}).

\section{\ours Benchmark Design}
\label{sec:method}

This section presents \ours as a controlled evaluation framework for spatial reasoning in \acp{vlm}.
Sec.~\ref{sec:210} defines the problem setup and evaluation protocol, specifying how agents interact with an environment and how success is measured.
Sec.~\ref{sec:220} introduces the controlled input conditions that form the core diagnostic design.
Sec.~\ref{sec:230} then describes task construction and verification, which together support consistent, deterministic, and scalable evaluation.

\begin{figure}[t]
  \centering
  \includegraphics[width=\textwidth]{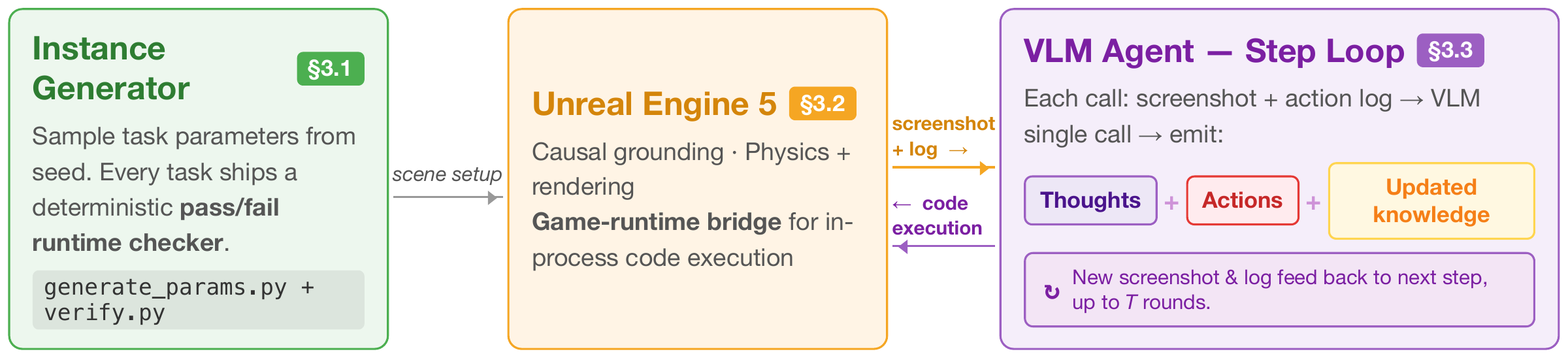}
  \caption{
    \textbf{\ours evaluation pipeline.}
    A Python generator maps a seed to task parameters and instantiates the scene in Unreal Engine~5 (rendering + game-runtime bridge for in-process code execution).
    A VLM agent runs a closed step loop: screenshot $+$ action log $\to$ thoughts / actions / updated knowledge $\to$ execute in UE5; the new screenshot and log feed back for up to $T$ rounds.
    A deterministic Python verifier checks pass / fail at runtime for \emph{every} task; some families additionally have a solvability certificate (Lean~4 or deterministic simulation, depending on the task; \S\ref{sec:230}).
  }
  \label{fig:pipeline}
\end{figure}

\subsection{Problem Setup and Evaluation Protocol}
\label{sec:210}

\ours is designed to evaluate whether a model can recover task-relevant spatial structure and use it to produce correct actions.
Each task is instantiated as an interactive 3D environment with a goal condition.
An agent interacts with the environment over multiple steps: it observes the current state, produces an executable action, and receives the resulting next state and observation. Depending on the task family, these actions may include executable code rather than only fixed control choices, allowing the agent to express procedures, parameters, or custom mathematical or physical programs directly. 
After the interaction budget is exhausted, or the task is terminated, the agent's outcome is evaluated by a deterministic verifier that returns a binary success or failure signal.
This defines a closed-loop evaluation protocol:
\emph{observation $\rightarrow$ action $\rightarrow$ environment transition $\rightarrow$ verification}.

A key feature of \ours is that the same underlying task instance can be evaluated under different information conditions while keeping the environment, objective, action interface, and verifier fixed.
Performance differences reflect the supplied observations, task information, and associated instructions under a common task and verifier.
Each task is constructed through a specification-to-environment pipeline (Fig.~\ref{fig:pipeline}), which we describe in Sec.~\ref{sec:230}.

\subsection{Controlled Input Conditions}
\label{sec:220}

The core diagnostic design of \ours is to evaluate the same task under input conditions that differ in the observations and task information supplied to the agent.
We use three conditions where applicable:
\begin{itemize}[leftmargin=*,topsep=0pt]
    \item \textbf{Visual (V).}
The agent receives rendered observations, action feedback, and task-specific metadata.
    \item \textbf{Structured (S).}
The agent receives structured task information without rendered observations. The supplied fields may describe object attributes, current state, target geometry, or physical parameters, depending on the task.
    \item \textbf{Combined (C).}
The agent receives rendered observations together with the task's structured information.
\end{itemize}
For a given task instance, the objective, environment dynamics, action interface, and verifier remain fixed.
Appendix~\ref{app:task_inputs} specifies the inputs for every task, including metadata already available in V and task-specific variants.
The S$-$V contrast measures the change in task performance under these specified inputs; its interpretation depends on whether the added information resolves object--identifier correspondence, supplies current state, or provides geometric and physical parameters.

\subsection{Task Construction and Verification}
\label{sec:230}

Each task family in \ours is defined by a specification that encodes both the task objective and its correctness criterion.
Tasks are constructed through a specification-to-instance pipeline, and evaluated using a runtime verifier that checks whether the agent's outcome satisfies the same criterion.

\paragraph{Task construction.}
Each task is defined by a symbolic specification that is independent of how it is rendered.
A specification defines a correctness predicate over valid solutions and constraints on task parameters (\eg, maze topology, object configuration, physical constants).
Task instances are generated parametrically from integer seeds and instantiated as interactive 3D environments.

\paragraph{Solvability check + runtime verification.}
For each benchmark instance, \ours runs a deterministic Python verifier at runtime that exits with a pass signal as soon as the task predicate is satisfied, or with fail once the agent reaches its interaction-budget limit---this runtime check is universal across all tasks. Whether an additional \emph{instance-level solvability certificate} is also needed depends on the task. For \textit{MazeNavFPS}, \textit{DropToTarget}, and \textit{CurvyBilliards}, the release includes a certificate of the appropriate form: Lean~4 per-seed theorems for discrete reachability (decided via \texttt{native\_decide}) or deterministic Slang/CUDA simulation for continuous physics. \textit{BoxFold} is solvable by construction because its flat layout is an unfolding of the cube; \textit{MovingShooter} has no separate solvability certificate. Tasks whose construction guarantees a solution---\textit{Billiards} and \textit{LightWar} (a parabolic trajectory trivially reaches any target within visible range), and \textit{SceneGrounding} and \textit{ObjectGrounding} (the target pose is itself the recovery objective)---do not require a separate certificate.
Thus all tasks are deterministically checked at runtime, and some task families additionally have a solvability certificate. Examples using Lean, Slang, and Python are provided in Appendix~\ref{app:verify_examples}.

\section{\ours Benchmark Overview}
\label{sec:benchmark}

\ours covers a compact set of interactive 3D task families chosen to expose the main diagnostic factors of spatial reasoning in \acp{vlm}: recovering task-relevant structure from visual input, reasoning over that structure once exposed, using feedback across interaction rounds, and acting through interfaces that range from step-wise control to executable or parameterized actions.
The task suite is designed to diagnose failures of perception, reasoning, multimodal input use, and interaction protocol.

At a coarse level, \ours includes two styles of agent--environment interaction.
In \textbf{interactive tasks}, the agent acts step by step and receives feedback after each action, incrementally building task-relevant structure through interaction.
In \textbf{computational tasks}, the agent proposes parameters or executable code whose effect is evaluated by the environment, often with feedback through simulation outputs such as videos or logs.
This distinction captures how the agent interfaces with the environment; the task families below capture the spatial reasoning challenges themselves.

\paragraph{Reporting tracks.}
Because the 11 tasks differ in observation modality, Sec.~\ref{sec:experiments} reports results in separate tracks rather than a single scalar leaderboard. The \emph{core information-condition track} contains tasks evaluated under V, S, and C on the same seed-generated task instances (V and C only for \textit{ObjectGrounding}); \emph{specialized tracks} cover tasks whose natural evaluation is visual-only or auxiliary (photoreal scene grounding, pose-assisted navigation); the \emph{harness-diagnostic track} holds task and information fixed while varying the action interface, exposing protocol confounds such as batch versus single-step control. Together, these tracks organize task-specific comparisons of perception, reasoning, multimodal input use, and interaction protocol while retaining all task families in the overall evaluation.

\begin{table}[t]
\centering
\caption{
\textbf{Task families in \ours.}
We organize 11 tasks into four families by interaction paradigm and the type of spatial reasoning required.
Family colours match the row banners of Table~\ref{tab:core_results}.
}
\label{tab:tasks}
\small
\setlength{\tabcolsep}{5pt}
\renewcommand{\arraystretch}{1.15}
\begin{tabular}{@{}p{2.7cm}p{5.2cm}p{4.8cm}@{}}
\toprule
\textbf{Family} & \textbf{Defining property} & \textbf{Tasks} \\
\midrule

\rowcolor{LightBlue!40}
\textbf{Interactive Control}
& Multi-round closed-loop interaction; the agent observes the scene, acts, and re-plans based on feedback. State persists across rounds.
& MazeNavFPS, Billiards, TunnelNav3D, PlaceOnBottom \\

\addlinespace[2pt]
\rowcolor{LightOrange!50}
\textbf{Trial-and-Error Prediction}
& The agent submits a structured candidate solution (action sequence, code, or numeric prediction) and receives binary pass/fail per attempt; iterates by trial and error.
& BoxFold, MovingShooter, LightWar \\

\addlinespace[2pt]
\rowcolor{LightGreen!60}
\textbf{Photoreal Grounding}
& Spatial-reasoning tasks instantiated in photo-realistic 3D scenes; the agent recovers a camera pose or localizes named objects from rendered views, given its current pose or marker coordinates.
& SceneGrounding, ObjectGrounding \\

\addlinespace[2pt]
\rowcolor{LightGray}
\textbf{Computational Physics}
& The agent either sets the launch angle and speed of a ball rolling on a curved surface (CurvyBilliards) or a surface-height function (DropToTarget); a deterministic GPU rollout checks the outcome, with up to five submit-and-check rounds.
& CurvyBilliards, DropToTarget \\

\bottomrule
\end{tabular}
\end{table}

\begin{figure}[!t]
\centering
\begin{tikzpicture}[
  thumb/.style 2 args = {draw=#1, line width=4.5pt, inner sep=0pt},
  tname/.style   = {font=\scriptsize, anchor=north, inner sep=1pt, yshift=-1pt}
]
\node[thumb={LightBlue}{}] (T_maze)   at ( 1.6, 5.6) {\includegraphics[width=2.9cm,height=2.175cm]{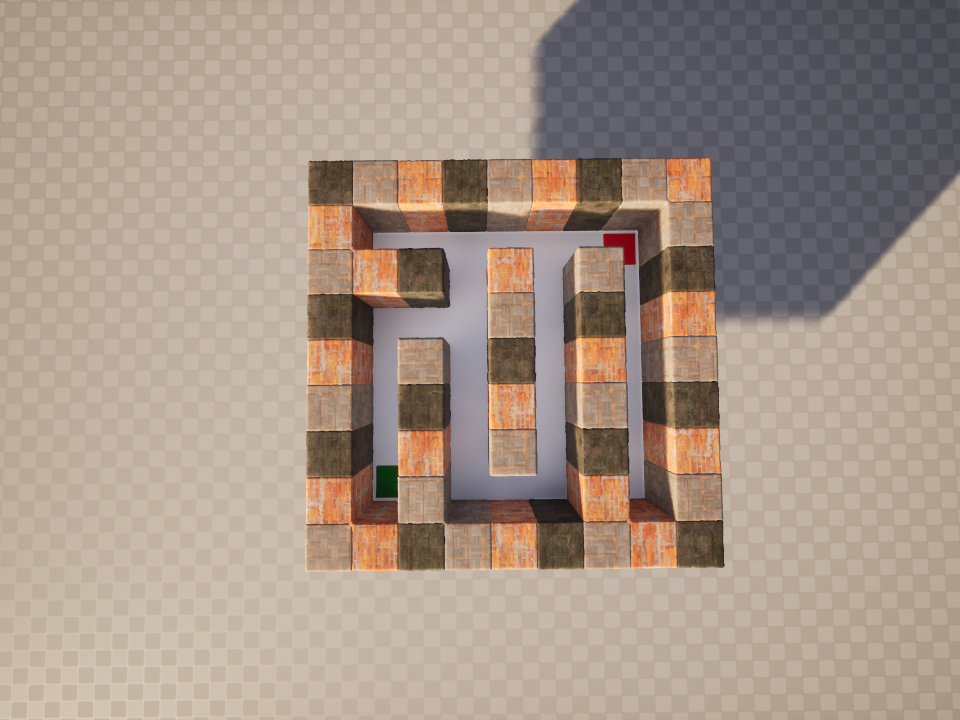}};
\node[tname, color=blue!70!black] at (T_maze.south) {\textbf{MazeNavFPS}};

\node[thumb={LightBlue}{}] (T_bill)   at ( 4.7, 5.6) {\includegraphics[width=2.9cm,height=2.175cm]{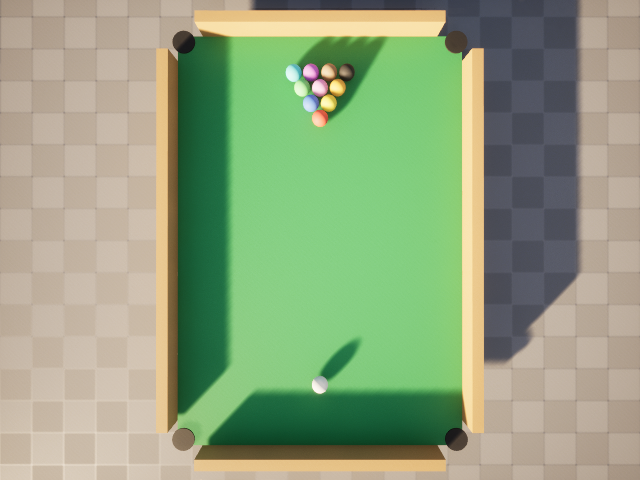}};
\node[tname, color=blue!70!black] at (T_bill.south) {\textbf{Billiards}};

\node[thumb={LightBlue}{}] (T_tunnel) at ( 7.8, 5.6) {\includegraphics[width=2.9cm,height=2.175cm]{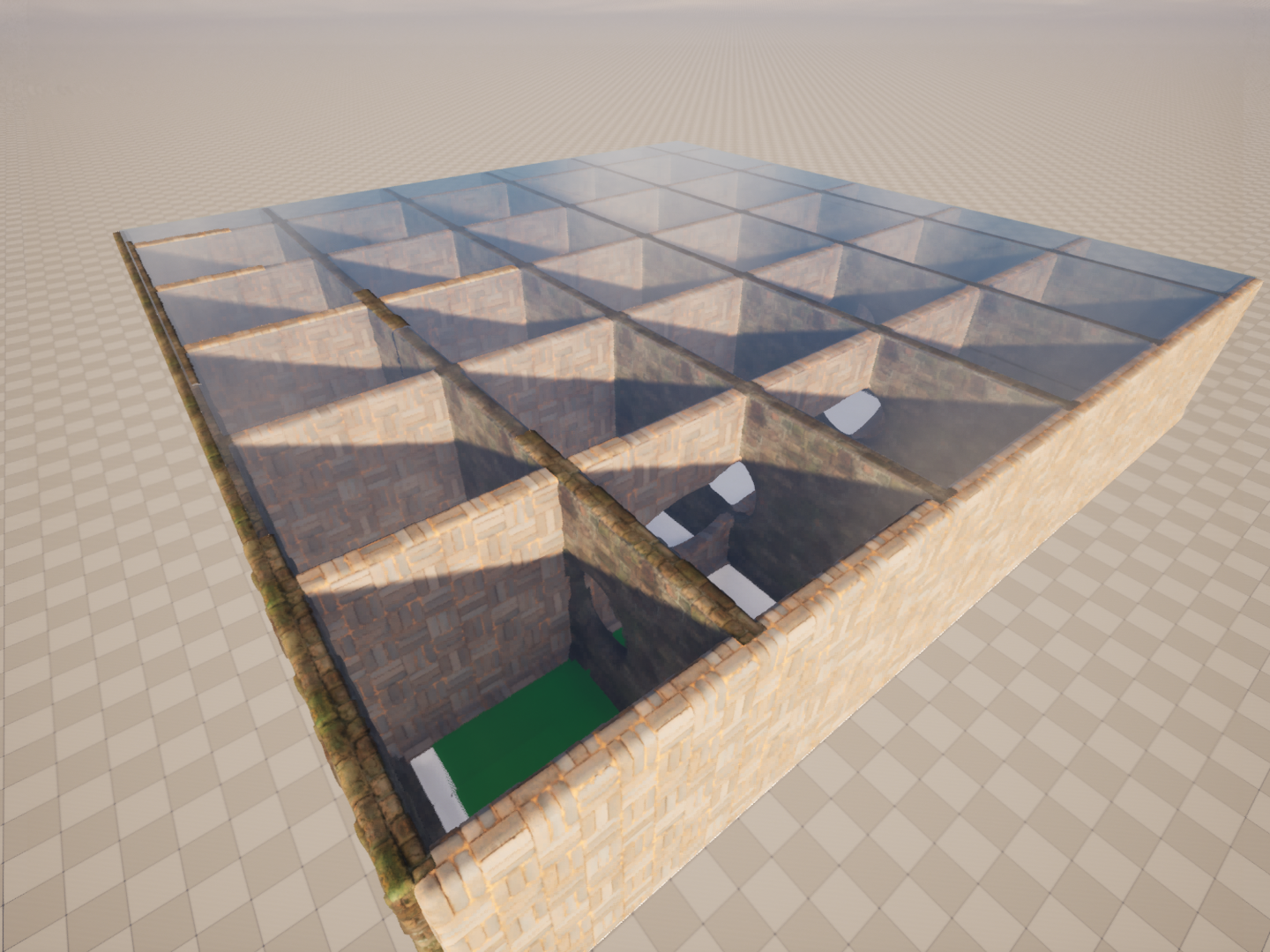}};
\node[tname, color=blue!70!black] at (T_tunnel.south) {\textbf{TunnelNav3D}};

\node[thumb={LightBlue}{}] (T_pob)    at (10.9, 5.6) {\includegraphics[width=2.9cm,height=2.175cm]{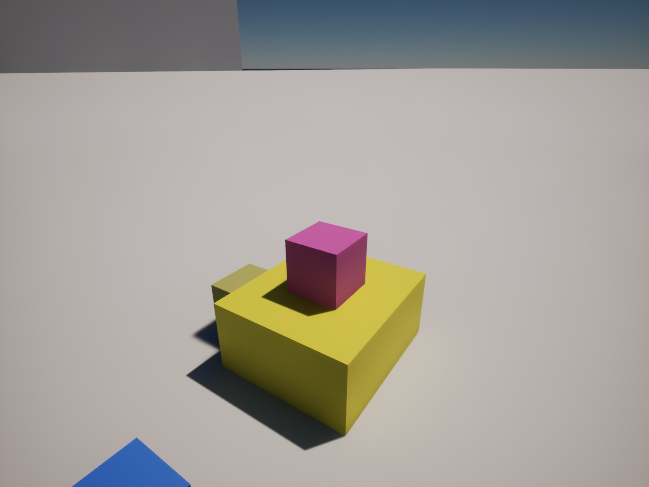}};
\node[tname, color=blue!70!black] at (T_pob.south) {\textbf{PlaceOnBottom}};

\node[thumb={LightOrange}{}] (T_box)   at ( 1.6, 3.0) {\includegraphics[width=2.9cm,height=2.175cm]{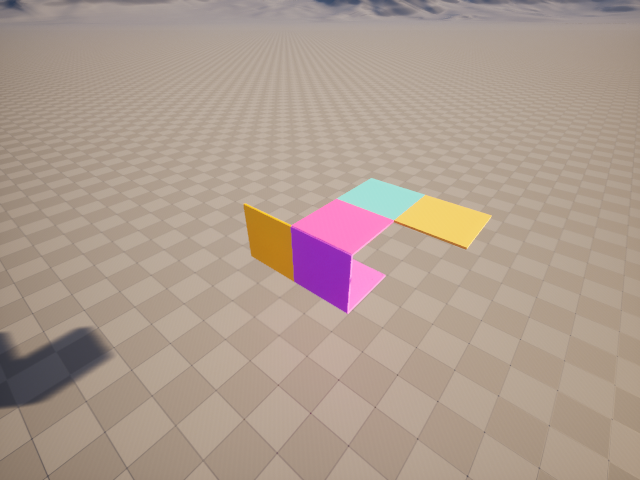}};
\node[tname, color=orange!80!black] at (T_box.south) {\textbf{BoxFold}};

\node[thumb={LightOrange}{}] (T_shoot) at ( 4.7, 3.0) {\includegraphics[width=2.9cm,height=2.175cm]{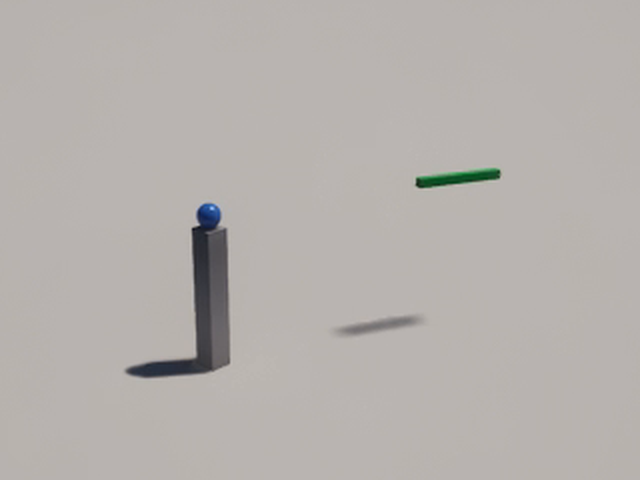}};
\node[tname, color=orange!80!black] at (T_shoot.south) {\textbf{MovingShooter}};

\node[thumb={LightOrange}{}] (T_light) at ( 7.8, 3.0) {\includegraphics[width=2.9cm,height=2.175cm]{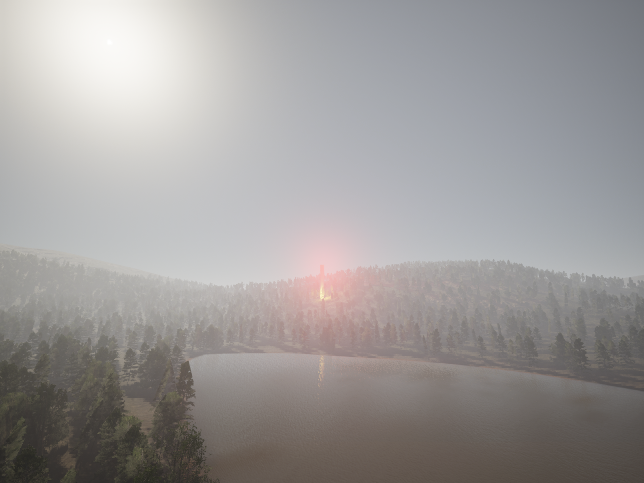}};
\node[tname, color=orange!80!black] at (T_light.south) {\textbf{LightWar}};

\node[thumb={LightGreen}{}] (T_scene)  at (10.9, 3.0) {\includegraphics[width=2.9cm,height=2.175cm]{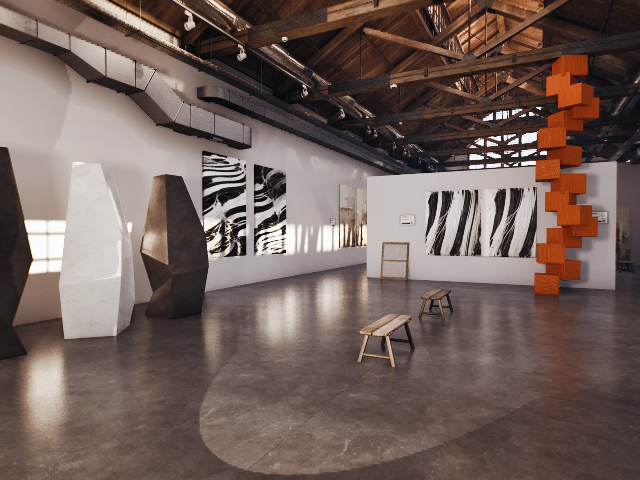}};
\node[tname, color=green!55!black] at (T_scene.south) {\textbf{SceneGrounding}};

\node[thumb={LightGreen}{}] (T_obj)    at ( 1.6, 0.4) {\includegraphics[width=2.9cm,height=2.175cm]{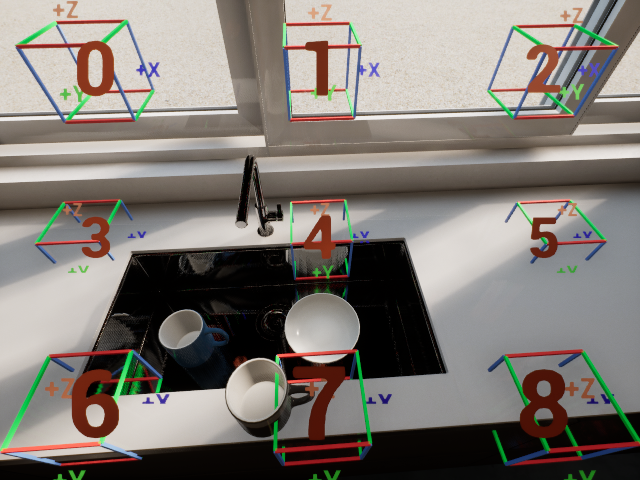}};
\node[tname, color=green!55!black] at (T_obj.south) {\textbf{ObjectGrounding}};

\node[thumb={black!50}{}] (T_cb)       at ( 4.7, 0.4) {\includegraphics[width=2.9cm,height=2.175cm]{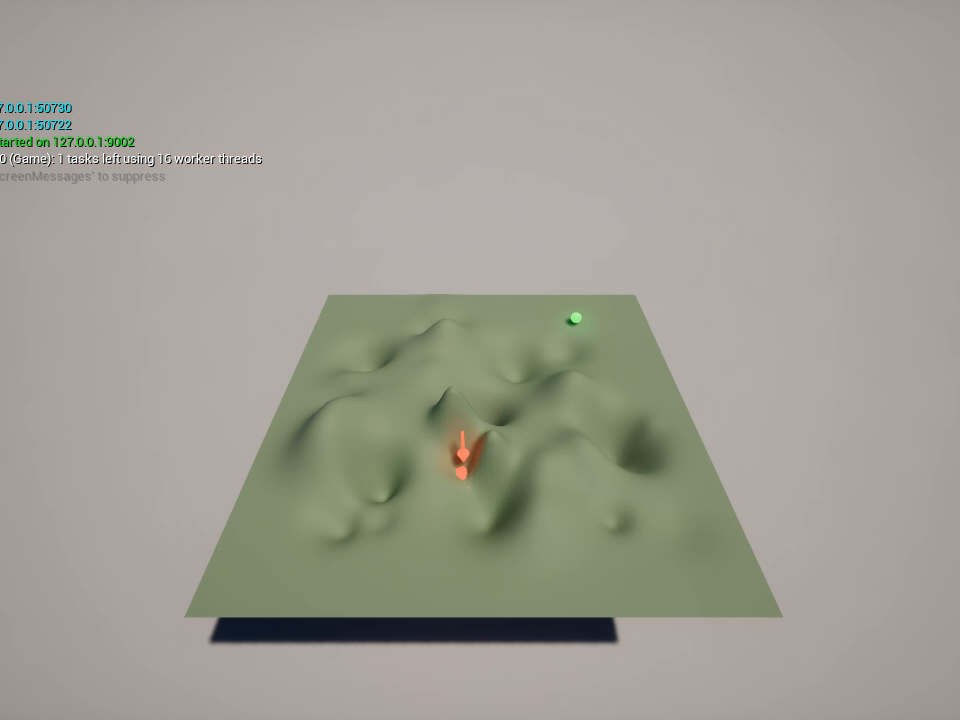}};
\node[tname, color=black!60] at (T_cb.south) {\textbf{CurvyBilliards}};

\node[thumb={black!50}{}] (T_d2t)      at ( 7.8, 0.4) {\includegraphics[width=2.9cm,height=2.175cm]{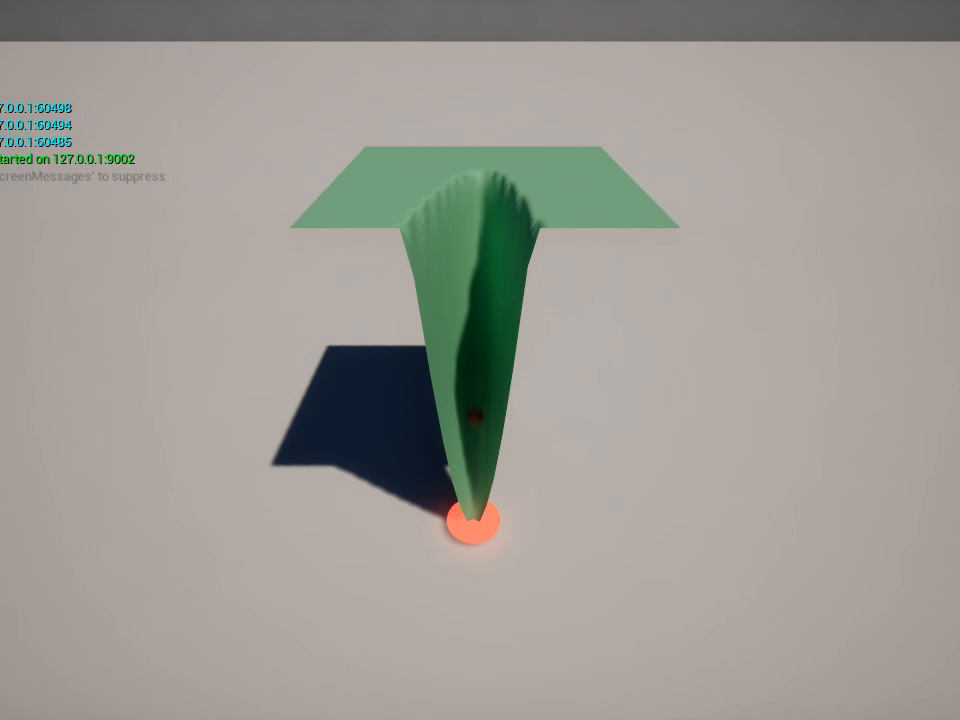}};
\node[tname, color=black!60] at (T_d2t.south) {\textbf{DropToTarget}};

\node[draw=black!40, rounded corners, fill=white, inner sep=4pt, font=\scriptsize, anchor=center] at (10.9, 0.4) {%
  \begin{tabular}{@{}r@{\;}l@{}}
    \textcolor{blue!70!black}{$\blacksquare$}   & Interactive Control \\
    \textcolor{orange!80!black}{$\blacksquare$} & Trial-and-Error \\
    \textcolor{green!55!black}{$\blacksquare$}  & Photoreal Grounding \\
    \textcolor{black!60}{$\blacksquare$}        & Computational Physics \\
  \end{tabular}};

\end{tikzpicture}
\caption{
\textbf{Overview of the 11 tasks in \ours, spanning four families.}
\textcolor{blue!70!black}{\textbf{Interactive Control}} (closed-loop, multi-round): \textit{MazeNavFPS} (first-person grid maze navigation), \textit{Billiards} (2D pool trajectory planning), \textit{TunnelNav3D} (curved-tunnel flythrough), \textit{PlaceOnBottom} (object stacking).
\textcolor{orange!80!black}{\textbf{Trial-and-Error Prediction}} (submit-and-check): \textit{BoxFold} (predict the folded shape), \textit{MovingShooter} (aim at a moving target), \textit{LightWar} (aim and launch a projectile at a distant beacon across a large outdoor terrain).
\textcolor{green!55!black}{\textbf{Photoreal Grounding}} (photo-realistic 3D scenes): \textit{SceneGrounding} (recover the camera pose of a given screenshot in two photoreal interior levels), \textit{ObjectGrounding} (localize named kitchen objects with 3D markers in a photoreal apartment scene).
\textcolor{black!60}{\textbf{Computational Physics}} (Slang/CUDA simulation): \textit{CurvyBilliards} (ball dynamics on a parametric surface), \textit{DropToTarget} (surface design for target landing).
Family colours match the row banners in Table~\ref{tab:tasks}; Sec.~\ref{sec:experiments} reports these tasks using separate core, specialized, and harness-diagnostic tracks.}
\label{fig:task_map}
\vspace{-18pt}
\end{figure}

\paragraph{Task Families \& Representative Tasks.}
Table~\ref{tab:tasks} summarizes the task families in \ours; each stresses a different part of the spatial reasoning pipeline (structure extraction, iterative search, geometric constraint satisfaction, parameter recovery for executable actions). Together they cover navigation, manipulation, photoreal grounding, geometric constraint satisfaction, and continuous-parameter recovery for executable actions. Each family contains many seed-parameterized instances drawn from a common specification, so models can be compared on matched tasks while reducing overfitting risk. Figure~\ref{fig:task_map} previews the 11 \ours tasks.

\section{Experiments}
\label{sec:experiments}

This section reports two diagnostic findings on \ours.
First, explicit task information often exposes reasoning competence that the visual condition does not reveal: models solve many tasks once object identity, current state, or target geometry is given, but fail when it must be read from rendered observations.
Second, the action interface and feedback harness can change outcomes under fixed tasks and fixed information.
Sec.~\ref{sec:setup} defines the setup; Sec.~\ref{sec:core_results} reports the core V/S/C benchmark; Sec.~\ref{sec:case_studies} examines harness effects. Specialized tracks and additional harness ablations are in the appendix.

\subsection{Experimental Setup}
\label{sec:setup}

\paragraph{Runtime and hardware.} We release Windows and Linux x86\_64 binaries that provide identical functionality and differ only in platform adaptation. The experiments in this paper were run on Windows workstations, with several Unreal Engine instances in parallel per GPU, and on Linux single-GPU cloud instances (mostly NVIDIA A40), each running one or two Unreal Engine instances. Hardware details and a multi-instance compatibility test are in Appendix~\ref{app:implementation}.

We evaluate \ours across three axes.
\begin{itemize}[leftmargin=*]
    \item \textbf{Models.} Each experiment uses up to twelve frontier VLMs from the Gemini, Claude, GPT, Qwen, Doubao, Kimi (Moonshot) and Grok families. Each table lists the models used for its experiment.
    \item \textbf{Information conditions.} V provides rendered observations together with task-specific metadata; S provides structured task information without images; C combines rendered observations with structured task information. Because agents act through code-level commands exposed by the \ours\ API rather than through keyboard and mouse input, often in world coordinates, V supplies the interface metadata these commands require (e.g., coordinate conventions, the agent's own pose and, for \textit{PlaceOnBottom}, object positions and sizes) and withholds the task-relevant information listed in Appendix~\ref{app:task_inputs}. Without this minimal metadata, visual reasoning has no frame in which to express world-coordinate commands (see Sec.~\ref{sec:core_results}). The appendix reports pixel-only controls that remove this metadata (Table~\ref{tab:pixel_only}); for example, only one of the seven model--task results exceeds 0.06. For a given task instance, the objective, verifier, and action interface remain fixed. This design evaluates how different ways of supplying task-relevant information affect task performance.
\item \textbf{Harness designs.} Each task also has an action interface matched to its objective, such as free forward/turn commands for \textit{MazeNavFPS}, \texttt{(angle, speed)} shots for \textit{Billiards}, launch \texttt{(yaw, pitch, speed)} for \textit{LightWar}, collision-aware step-and-turn or walk-and-set-view commands for indoor view recovery in \textit{SceneGrounding}, and a Python surface function or numerical launch parameters evaluated by GPU simulation for the computational tasks.
The full action-space catalogue is provided in the appendix.
\end{itemize}

The V/S/C contrast makes \ours diagnostic by showing how explicit task information and visual observations affect performance under the input configurations specified in Appendix~\ref{app:task_inputs}.
Not every task family admits a clean V/S/C triple on the same instance.
Photoreal scene grounding and pose-assisted navigation are reported as specialized tracks because their observation modalities leave one or more channels ill-defined, and harness ablations are reported separately because they vary the action interface rather than the information channel.
For harness-diagnostic variants we use shorthand: VP $=$ rendered observation $+$ pose/position history; RP $=$ raycast $+$ pose/position without rendering; VRP $=$ rendered observation $+$ raycast $+$ pose/position; V+REPL $=$ visual input $+$ a self-feedback loop that renders the agent's proposed action before commitment.
These are auxiliary variants, not core V/S/C conditions.

All tasks use a knowledge-augmented ReAct agent over seed-generated instances with a task-specific budget, set either as a number of model-call rounds or as a number of action moves (detailed in the appendix). We report Pass@1 averaged over seeds; for \textit{Billiards}, Pass@1 is the mean fraction of balls pocketed. The main text reports the canonical result slices for each reporting track while preserving condition semantics. Sample counts and uncertainty estimates are in the appendix.

\subsection{Models Succeed with Explicit Structure but Struggle to Obtain It from Visual Input}
\label{sec:core_results}

\begin{table}[!t]
\centering
\caption{
\textbf{Core V/S/C information-condition results.}
We include tasks for which visual (V), structured-only (S), and combined (C) conditions are defined on seed-generated task instances with the same action interface and verifier; \textit{ObjectGrounding} has no structured-only condition and is reported with V and C.
V provides rendered observations and task-specific metadata describing the scene setup and action interface, while retaining task-specific demands for visual identification or state estimation.
S provides explicit information such as object attributes, current or target coordinates, and physical parameters without images.
The S$-$V gap measures the resulting change in task performance (Appendix~\ref{app:task_inputs}).
The C$-\max$(V,S) column measures whether simply combining rendered and structured inputs improves over the better single channel.
Positive S$-$V indicates a benefit from explicit task information, which can reduce demands on visual identification and state estimation; negative C$-\max$(V,S) means that the combined condition, including its task-specific instructions, underperforms the better single channel, and does not by itself isolate a fusion deficit.
}
\label{tab:core_results}
\small
\setlength{\tabcolsep}{3.5pt}
\begin{tabular}{@{}llrrrrr@{}}
\toprule
\textbf{Task} & \textbf{Model} & \textbf{V} & \textbf{S} & \textbf{C} & \textbf{S$-$V} & \textbf{C$-\max$(V,S)} \\
\midrule
\multicolumn{7}{@{}l}{\textit{Interactive Control}} \\
\midrule
\multirow{3}{*}{Billiards$^{*}$}
& Gemini-3-Pro & 0.33 & 0.67 & 0.53 & +0.33 & -0.13 \\
& Sonnet       & 0.20 & 0.47 & 0.53 & +0.27 & +0.07 \\
& GPT-5.2      & 0.13 & 0.40 & 0.27 & +0.27 & -0.13 \\
\addlinespace
\multirow{2}{*}{PlaceOnBottom}
& Gemini-3.1-Pro & 0.85 & 1.00 & 1.00 & +0.15 & 0.00 \\
& Opus-4.5 & 0.35 & 1.00 & 1.00 & +0.65 & 0.00 \\
\midrule
\multicolumn{7}{@{}l}{\textit{Trial-and-Error Prediction}} \\
\midrule
\multirow{3}{*}{BoxFold}
& Gemini-3.1-Pro & 0.60 & 1.00 & 1.00 & +0.40 & 0.00 \\
& Flash-2.5 & 0.00 & 0.33 & 0.33 & +0.33 & 0.00 \\
& Sonnet-4.6 & 0.02 & 0.40 & 0.38 & +0.38 & -0.02 \\
\addlinespace
MovingShooter
& Gemini-3.1-Pro & 0.00 & 0.30 & 0.14 & +0.30 & -0.16 \\
\addlinespace
\multirow{3}{*}{LightWar$^{\ddagger}$}
& Gemini-3.1-Pro & 0.63 & 1.00 & 1.00 & +0.37 &  0.00 \\
& Sonnet-4.6     & 0.04 & 1.00 & 1.00 & +0.96 &  0.00 \\
& GPT-5.2        & 0.00 & 1.00 & 1.00 & +1.00 &  0.00 \\
\midrule
\multicolumn{7}{@{}l}{\textit{Photoreal Grounding}} \\
\midrule
\multirow{3}{*}{ObjectGrounding$^{\sharp}$}
& Gemini-3.1-Pro & 0.69 & -- & 0.90 &  -- & +0.21 \\
& GPT-5.2        & 0.32 & -- & 0.57 &  -- & +0.24 \\
& Sonnet         & 0.21 & -- & 0.27 &  -- & +0.06 \\
\midrule
\multicolumn{7}{@{}l}{\textit{Computational Physics}} \\
\midrule
\multirow{4}{*}{DropToTarget$^{\dagger}$}
& Gemini-3.1-Pro & 0.11 & 1.00 & 0.89 & +0.89 & -0.11 \\
& GPT-6.1-Sol    & 0.44 & 1.00 & 1.00 & +0.56 &  0.00 \\
& Sonnet-4.6     & 0.06 & 0.67 & 0.67 & +0.61 &  0.00 \\
& Gemini-3-Flash & 0.11 & 0.56 & 0.50 & +0.44 & -0.06 \\
\addlinespace
\multirow{3}{*}{CurvyBilliards$^{\dagger}$}
& Opus-4.5       & 0.00 & 0.11 & 0.11 & +0.11 &  0.00 \\
& Gemini-3.1-Pro & 0.17 & 0.28 & 0.00 & +0.11 & -0.28 \\
& GPT-6.1-Sol    & 0.28 & 0.28 & 0.11 &  0.00 & -0.17 \\
\bottomrule
\end{tabular}
\vspace{2pt}
\begin{minipage}{0.96\linewidth}
\footnotesize
For \textit{PlaceOnBottom}, \textit{BoxFold}, and Gemini-3.1-Pro on \textit{MovingShooter}, each condition is averaged over all valid seeds; per-condition sample counts and matched-seed analyses are reported in Appendix~\ref{app:supplementary_experiments}.
$^{*}$For \textit{Billiards}, Pass@1 is the mean fraction of balls pocketed (standard preset: 5 balls $\times$ 5 rounds, 3 seeds); three of nine models are shown, with all nine in Table~\ref{tab:supp_billiards}.
$^{\dagger}$Each computational-physics score averages six equal-weight units per condition: three seeds and three scene configurations (three runs each).
$^{\ddagger}$\textit{LightWar}: matched V/S/C results; sample accounting is in Appendix~\ref{app:harness_lightwar}. The V$+$Telemetry harness ablation in Appendix~\ref{app:harness_lightwar} provides the detailed harness analysis.
$^{\sharp}$For \textit{ObjectGrounding}, both V and C use rendered scenes with the camera pose and a lookup table from marker index to world coordinates; C renders each of the nine markers' index labels large and unoccluded, whereas V uses the original small labels inside the wireframe. A structured-only condition is not defined because the rendered scene is required to disambiguate semantic targets (e.g., ``the wooden spoon'' vs.\ ``the black spoon''). Pass@1 over $\sim$30 seeds per cell; the full 12-model panel and the contrasting single-marker harness are in Appendix~\ref{app:harness_object_ground}.
Sample counts and uncertainty estimates are reported in the appendix.
\end{minipage}
\vspace{-12pt}
\end{table}

Table~\ref{tab:core_results} shows that structured information is often decisive.
On \textit{BoxFold} and \textit{PlaceOnBottom}, the same models perform far better under S than under V. In \textit{BoxFold}, V already supplies the unfolded layout and hinge tree, while S adds the current and target face-centre coordinates, so the gap reflects reading the current fold state from images.
For \textit{PlaceOnBottom}, V already lists the current position and size of every object each round (Appendix~\ref{app:task_inputs}), so its S$-$V gap reflects identifying which object is which given that geometry. A pixel-only control that also removes these coordinates (images and anonymous identifiers only) lowers Pass@1 on the same seeds from 0.85 to 0.10 for Gemini-3.1-Pro and from 0.34 to 0.00 for Opus-4.5; because \texttt{move\_to} still takes absolute world coordinates and no coordinate reference is given, this control mainly tests recovering a world frame from images. A pixel-only \textit{MazeNavFPS} condition with egocentric actions, by contrast, remains informative (Appendix~\ref{app:maze_results_section}).
Large gaps occur in \textit{BoxFold}, where S exceeds V by +0.33 to +0.40.
In the computational-physics tasks, S exceeds V on \textit{DropToTarget} for all four reported models (+0.44 to +0.89), whereas \textit{CurvyBilliards} remains difficult under all three conditions (S$-$V $\le +0.11$).
\textit{ObjectGrounding} extends the comparison to photoreal target localization, where a structured-only condition cannot be defined. Both conditions provide the marker-to-world-coordinate table; making the nine markers' index labels readable (C) raises Pass@1 for the three featured models (9 of 12 in the full panel), with $C-V$ as high as $+0.24$ for GPT-5.2 and $+0.21$ for Gemini-3.1-Pro, showing that how marker information is presented matters in a task that still requires visual target identification.

The last column, C$-\max$(V,S), shows that giving the model both images and structured information does not reliably beat the better single input.
For \textit{BoxFold}, \textit{PlaceOnBottom}, and \textit{MovingShooter}, C matches or falls below S, suggesting that the combined condition does not necessarily improve performance over structured input alone.
For \textit{Billiards}, C is best for several of the nine models but worse than S or V for others (Table~\ref{tab:supp_billiards}).
The main diagnostic message is that V/S/C comparisons reveal task-specific benefits and difficulties in using visual and structured inputs, helping identify where closer analysis of perception, state estimation, or multimodal integration is needed.
LightWar illustrates the role of instructions: retaining images and structured information but changing the instruction raises GPT-4o's success rate from 0.00 to 0.60 on ten matched seeds (Appendix~\ref{app:supp_combined_example}). Thus the original C$-$S gap does not isolate a fusion effect.

\subsection{Changing Only the Harness Can Flip the Outcome}
\label{sec:case_studies}

We now isolate the harness axis: under fixed task instance, model, and information condition, we vary only the harness (action interface, feedback, or auxiliary input channels) and ask whether this change alone can flip outcomes. \ours\ exposes four conceptually distinct sub-axes --- \textbf{action span}, \textbf{intra-round feedback}, \textbf{input-channel set}, and \textbf{control protocol} --- which we unpack on mazes, billiards, and \textit{LightWar} below; the tunnel control-protocol axis is in Appendix~\ref{app:harness_extra}. Action span specifies how many primitive actions are committed per model call, whereas control protocol specifies the action vocabulary and how interaction state is maintained. The comparisons include each protocol's action allowances and feedback schedule; for grid mazes, we compare protocols under two budget definitions, an equal number of model calls and an equal number of primitive moves, and for curved tunnels under an equal number of primitive moves.

\begin{table}[t]
\centering
\caption{
\textbf{Grid mazes under an equal primitive-move budget: batch outperforms single-step control under both information variants.}
Pass@1 over 30 mazes per cell (grids 3, 4, 5 $\times$ 10 seeds). Batch and single share a budget of 125, 170 and 210 primitive moves for $3{\times}3$, $4{\times}4$ and $5{\times}5$ grids; the batch prompt asks for 3--6 commands per model call and single sends one. B-only:S-only counts the mazes solved only by batch and only by single, which shows whether batch wins on the same mazes or the two protocols simply solve different ones.
VP = rendered observation + position history; VRP = VP + a 4-direction raycast scan.
These are auxiliary harness variants, not core V/S/C conditions.
}
\label{tab:case_maze}
\small
\setlength{\tabcolsep}{3pt}
\begin{tabular}{@{}lcccccc@{}}
\toprule
& \multicolumn{3}{c}{\textbf{VP}} & \multicolumn{3}{c}{\textbf{VRP}} \\
\cmidrule(lr){2-4}\cmidrule(lr){5-7}
\textbf{Model} & \textbf{Batch} & \textbf{Single} & \textbf{B-only:S-only} & \textbf{Batch} & \textbf{Single} & \textbf{B-only:S-only} \\
\midrule
Gemini-3.1-Pro & 0.93 & 0.57 & 12:1 & 0.97 & 0.57 & 12:0 \\
Sonnet-4.6 & 0.83 & 0.70 & 5:1 & 0.73 & 0.37 & 11:0 \\
GPT-5.2 & 0.67 & 0.43 & 8:1 & 0.60 & 0.43 & 6:1 \\
GPT-4.1 & 0.47 & 0.33 & 7:3 & 0.67 & 0.33 & 10:0 \\
\midrule
Pooled & 0.72 & 0.51 & 32:6 & 0.74 & 0.42 & 39:1 \\
\bottomrule
\end{tabular}
\vspace{2pt}
\begin{minipage}{0.96\linewidth}
\footnotesize
Per-grid results, the equal model-call comparison, and a response-parsing check are in Appendix~\ref{app:harness_extra}.
\end{minipage}
\vspace{-15pt}
\end{table}

\paragraph{Grid mazes: batch protocols outperform single-step control under both budget definitions.}
Table~\ref{tab:case_maze} compares batch and single-step control when both share the same primitive-move budget. Batch outperforms single under VP (pooled 0.72 vs.\ 0.51; 32 vs.\ 6 mazes solved by only one protocol) and under VRP (0.74 vs.\ 0.42; 39 vs.\ 1), and batch uses fewer primitive moves than single (101.13 vs.\ 120.31 under VP; 95.72 vs.\ 105.72 under VRP). The size of the gain varies by model; Appendix~\ref{app:harness_extra} also reports a response-parsing check for GPT-4.1 under VRP. Under an equal budget of 50 model calls, where batch can execute more primitive moves than single, batch also exceeds single for every reported model and variant (Appendix~\ref{app:harness_extra}); in that comparison, raycast does not improve uniformly, helping GPT-4.1 under single-step control but hurting GPT-5.2 and Qwen3-235B.
A complementary control-protocol study on \textit{TunnelNav3D} gives every protocol the same budget of 80 primitive moves: active aim-and-fly control reaches 0.77 Pass@1, versus 0.18 for batch and 0.29 for single-step control, and is best or tied for best for every model (Appendix~\ref{app:harness_tunnel}).

\paragraph{Billiards: visual self-feedback flips outcome without revealing new state.}
The V+REPL harness lets the agent preview its own proposed shot as a rendered aim-line and revise it up to three times before firing, but reveals no additional hidden information --- action vocabulary, world state, and information channel are unchanged. On the hard preset (three seeds per model), this self-feedback raises Pass@1 for 9 of 11 models, including Flash $0.40 \to 0.80$ and Sonnet $0.33 \to 0.67$ (Appendix~\ref{app:harness_aimline}). Like action span, this feedback changes success rates substantially, but for a different reason; the harness therefore affects results in more than one way.

\paragraph{LightWar: input-channel set alters outcomes without changing scene state.}
A complementary axis adds one auxiliary input channel (V vs.\ V+Telemetry, which also reports each shot's miss distance) on outdoor projectile aim. With telemetry the 12-model panel (three seeds per model) reaches 0.17 Pass@1 vs.\ 0.06 without; the gap is driven by four mid-tier models (Sonnet, Opus, Kimi, Qwen3-VL-30B) that pass once with telemetry and never without, while Gemini~3.1~Pro is unaffected (Appendix~\ref{app:harness_lightwar}).

\paragraph{Cross-axis pattern.}
Harness effects vary across tasks and models: action protocols, feedback loops, or input channels can shift Pass@1 from 0.00 to $\sim$0.80 in the reported comparisons. These differences show why evaluation must specify inputs, instructions, action allowances, and feedback schedules. They identify configurations that help a model use task information, without uniquely attributing the improvement to perception, fusion, or planning.

\section{Conclusion}
\label{sec:conclusion}
\ours is a Visual SWE-Bench for spatial agents: interactive 3D environments, executable actions, and deterministic verification support task-specific comparisons of perception, reasoning, and interaction. Matched V/S/C inputs reveal a recurring visual-to-structure gap, while action protocols and feedback can substantially affect outcomes even when protocols share the same primitive-move budget, making the harness a first-class evaluation variable.
\textbf{Limitations.} The benchmark is diagnostic, not exhaustive: image-interactive and computational-physics tasks form the main panel, with specialized tracks and harness ablations on selected families.
\section*{Acknowledgments and Disclosure of Funding}
We thank 2077 AI and Abaka AI for API credits. Bangya Liu and Zihan Wang are employees of Abaka AI. The other authors declare no competing interests.

\clearpage
\bibliographystyle{abbrvnat}
\bibliography{references}

\newpage
\appendix
\renewcommand{\thesection}{\Alph{section}}
\begin{center}
\vspace*{0.5em}
\rule{0.92\textwidth}{0.6pt}\\
\vspace{0.6em}
{\LARGE\bfseries Appendix\par}
\vspace{0.3em}
{\large\itshape Supplementary Material for VeriWorld\par}
\vspace{0.4em}
\rule{0.92\textwidth}{0.6pt}
\end{center}
\vspace{0.8em}

\section{Per-Task Catalogue}
\label{app:task_catalogue}

This section gives a one-row-per-task summary of all eleven \ours tasks across the four families introduced in Sec.~\ref{sec:benchmark}.
Each row reports the rendered scene used, the agent's objective, its per-round action space, and the deterministic verifier that decides Pass/Fail.
Thumbnails reproduce the same task art shown in Fig.~\ref{fig:task_map}; we re-include them here so the table can stand alone as a reproducibility reference.

\begingroup
\small
\setlength{\tabcolsep}{3pt}
\renewcommand{\arraystretch}{1.15}
\begin{longtable}{@{}c p{2.2cm} p{3.9cm} p{4.2cm}@{}}
\caption{
\textbf{\ours per-task catalogue.}
Eleven tasks across four families.
$T$ denotes the per-task interaction-budget (max rounds before the verifier returns FAIL).
For \textit{Billiards}, Pass@1 is the mean fraction of balls pocketed; all other tasks are binary.
Action-space and verifier columns are truncated; see the corresponding harness module under \texttt{veriworld/benchmark/} for the exact signature.
}
\label{tab:per_task_catalogue} \\
\toprule
\textbf{Thumbnail} & \textbf{Task / Family} & \textbf{Setting \& Objective} & \textbf{Action / Verifier} \\
\midrule
\endfirsthead
\multicolumn{4}{c}{\tablename\ \thetable\ (continued)} \\
\toprule
\textbf{Thumbnail} & \textbf{Task / Family} & \textbf{Setting \& Objective} & \textbf{Action / Verifier} \\
\midrule
\endhead
\bottomrule
\endfoot
\multicolumn{4}{@{}l}{\textit{Interactive Control (closed-loop, multi-round)}} \\*
\midrule
\includegraphics[width=1.4cm,height=1.1cm]{figures/cropped/maze_navfps.png} &
\textbf{MazeNavFPS} \newline {\footnotesize Interactive Control} &
First-person view of an $N{\times}N$ corridor maze; agent must reach a hidden goal cell. &
{\footnotesize Action: free \texttt{forward}(distance) / \texttt{turn}(degrees) commands, batched or single. Verifier: reach goal within 30--50 model calls (equal-move study: 125--210 primitive moves).} \\
\addlinespace[2pt]
\includegraphics[width=1.4cm,height=1.1cm]{figures/cropped/billiards.png} &
\textbf{Billiards} \newline {\footnotesize Interactive Control} &
Top-down 2D pool table with cue ball, 5--10 target balls, 4 corner pockets; multi-round trajectory planning. &
{\footnotesize Action: $(\text{angle}^\circ,\, \text{speed cm/s})$ per round. Verifier: target balls pocketed across $T{=}5$ (standard) or 10 (hard) rounds. } \\
\addlinespace[2pt]
\includegraphics[width=1.4cm,height=1.1cm]{figures/cropped/tunnel_nav3d.png} &
\textbf{TunnelNav3D} \newline {\footnotesize Interactive Control} &
First-person flythrough of a curved 3D tunnel with one circular exit hole at the far end. &
{\footnotesize Action: per-step (yaw, pitch, forward distance). Verifier: a logged position comes within two cells (400\,cm) of the exit within 80 primitive moves (information-channel runs: 30 model calls). } \\
\addlinespace[2pt]
\includegraphics[width=1.4cm,height=1.1cm]{figures/cropped/place_on_bottom.png} &
\textbf{PlaceOnBottom} \newline {\footnotesize Interactive Control} &
Tabletop scene with one target cube, one platform, and 2--8 distractor cubes/spheres of varied size and colour. &
{\footnotesize Action: \texttt{probe}, \texttt{move\_to}$(x, y, z)$, or \texttt{stop} per round. Verifier: after physics settles, target within 5\,cm of its resting height and inside the platform footprint, $T{=}8$ rounds. } \\
\newpage
\multicolumn{4}{@{}l}{\textit{Trial-and-Error Prediction (submit-and-check, multi-round)}} \\*
\midrule
\includegraphics[width=1.4cm,height=1.1cm]{figures/cropped/box_fold.png} &
\textbf{BoxFold} \newline {\footnotesize Trial \& Error} &
Six face-slabs laid flat as one of 9 distinct hexomino unfoldings (T-cross, +, strip-4, Z, 2-2-2 staircase, etc.). &
{\footnotesize Action: $(\text{hinge\_id}, \text{angle}^\circ)$ per round (rigid subtree rotation). Verifier: 6 face-centres within $\sim$8\,cm of closed-cube positions, $T{=}15$ rounds. } \\
\addlinespace[2pt]
\includegraphics[width=1.4cm,height=1.1cm]{figures/cropped/moving_shooter.png} &
\textbf{MovingShooter} \newline {\footnotesize Trial \& Error} &
Blue ball orbits a gray pillar (XY circle + Z oscillation); a green arm orbits at larger radius and tracks the target until fire. &
{\footnotesize Action: $(\text{fire\_time s},\, \allowbreak \text{aim\_yaw}^\circ,\, \allowbreak \text{aim\_pitch}^\circ)$. Verifier: projectile within hit\_radius of target during sim; $T{=}5$ submissions. } \\
\addlinespace[2pt]
\includegraphics[width=1.4cm,height=1.1cm]{figures/cropped/lightwar.png} &
\textbf{LightWar} \newline {\footnotesize Trial \& Error} &
Outdoor MAEBeechForest scene; agent fires a parabolic projectile from a fixed launcher at a distant target. &
{\footnotesize Action: launch $(\text{yaw}^\circ,\, \text{pitch}^\circ,\, \text{speed cm/s})$ + optional V$+$Telemetry channel. Verifier: landing point within 30\,m of the beacon apex; $T{=}8$ rounds. } \\
\midrule
\multicolumn{4}{@{}l}{\textit{Photoreal Grounding (photo-realistic interior scenes)}} \\*
\midrule
\includegraphics[width=1.4cm,height=1.1cm]{figures/cropped/scene_grounding.png} &
\textbf{SceneGrounding} \newline {\footnotesize Photoreal} &
Two photoreal interior levels (living room, AIUE); agent is given one screenshot from an unknown camera pose and must navigate to that pose. &
{\footnotesize Action: collision-aware \texttt{forward}/\texttt{backward} steps with relative \texttt{turn}, or \texttt{walk} with absolute \texttt{set\_view}. Verifier: final pose within $\le$100\,cm xy and $\le$10$^\circ$ yaw (and pitch, step/turn only) of GT in $T{=}15$ rounds. } \\
\addlinespace[2pt]
\includegraphics[width=1.4cm,height=1.1cm]{figures/cropped/object_grounding.png} &
\textbf{Object\allowbreak Grounding} \newline {\footnotesize Photoreal} &
Kitchen scene with $\sim$10 named objects (mug, pan, spoon variants); offcenter camera; target named via NL phrase. &
{\footnotesize Action: place / move colored bbox markers; nominate the marker closest to the target. Verifier: claimed-closest marker within tol. of GT object pose, $T{=}10$ rounds. } \\
\midrule
\multicolumn{4}{@{}l}{\textit{Computational Physics (Slang / CUDA simulation)}} \\*
\midrule
\includegraphics[width=1.4cm,height=1.1cm]{figures/cropped/curvy_billiards.png} &
\textbf{CurvyBilliards} \newline {\footnotesize Comp.\ Physics} &
A ball rolls on a parametric curved surface (Gaussian bumps, saddles); agent must hit a target patch from a fixed launch point. &
{\footnotesize Action: launch angle and speed. Verifier: deterministic GPU rollout checks whether the ball settles inside the target region. } \\
\addlinespace[2pt]
\includegraphics[width=1.4cm,height=1.1cm]{figures/cropped/drop_to_target.png} &
\textbf{DropToTarget} \newline {\footnotesize Comp.\ Physics} &
A ball falls onto a surface designed by the agent and must first touch the ground inside a target circle. &
{\footnotesize Action: a Python \texttt{surface\_height(x,y)} function. Verifier: deterministic GPU rollout checks whether the ball centre lies within the target radius at first ground contact. } \\
\end{longtable}
\endgroup

The four families partition tasks by interaction style (closed-loop vs.\ submit-and-check) and by what the agent must reason over (action sequences, parametric predictions, photoreal landmarks, simulation parameters).
This split is what motivates the reporting-tracks structure used in Sec.~\ref{sec:experiments}: tasks within a family share enough machinery that V/S/C contrasts mean the same thing across rows, while across-family comparisons are reported as separate tracks rather than averaged into a single leaderboard.
Sample sizes vary across tasks: \textit{ObjectGrounding} and \textit{PlaceOnBottom} include 12-model sweeps; \textit{TunnelNav3D} uses a single tunnel geometry, with 286 scored episodes in the equal-move protocol study and 191 (model $\times$ harness-config) cells behind Table~\ref{tab:specialized_results}; the computational-physics tasks (\textit{CurvyBilliards}, \textit{DropToTarget}) use three seeds plus three scene configurations per model and condition.

\subsection{Per-Task Input Specifications}
\label{app:task_inputs}

The input conditions below specify what the model receives, rather than all state accessible to the simulator. Task objectives, action formats, and interaction limits are given in every condition. V can include numerical metadata; S supplies the fields listed for that task, rather than a uniformly complete scene description. Navigation channels and photoreal marker variants retain their own meanings.

\begingroup
\footnotesize
\setlength{\tabcolsep}{3pt}
\renewcommand{\arraystretch}{1.12}
\begin{longtable}{@{}>{\raggedright\arraybackslash}p{1.65cm}>{\raggedright\arraybackslash}p{3.65cm}>{\raggedright\arraybackslash}p{4.15cm}>{\raggedright\arraybackslash}p{3.65cm}@{}}
\toprule
\textbf{Task} & \textbf{Visual observations and supplied metadata} & \textbf{Structured or combined inputs} & \textbf{Interpretation of the contrast} \\
\midrule
\endfirsthead
\toprule
\textbf{Task} & \textbf{Visual observations and supplied metadata} & \textbf{Structured or combined inputs} & \textbf{Interpretation of the contrast} \\
\midrule
\endhead
\bottomrule
\endfoot
PlaceOn\-Bottom &
V: screenshot plus anonymous object identifiers, current positions and sizes. The goal and placement rule are stated, but per-object colour and shape attributes are omitted. &
S: positions and sizes with per-object colour and shape attributes, without images. C: the structured listing plus the screenshot. &
Associating visible objects with identifiers, then using supplied geometry for placement. V already contains numerical geometry. \\
\addlinespace[4pt]
BoxFold &
V: unfolded face layout, hinge tree and dimensions, with images of the current folded state. &
S: hinge tree, dimensions, current face-centre coordinates and explicit target face-centre coordinates, without images. C: the layout, images and structured current/target coordinates. &
Use of explicit current and target state versus interpreting the rendered fold; the unfolded topology is already supplied in V. \\
\addlinespace[4pt]
Moving\-Shooter &
V: time-ordered replay frames, firing-time/yaw/pitch action definitions, previous actions and HIT/MISS feedback. No motion equations are supplied. &
S: target, shooter and projectile equations, action history and simulation logs, without images. C adds replay frames. The equations use a rotated frame whose offset from the action's world frame is not explicitly disclosed. &
Motion interpretation and use of equations with a remaining coordinate-frame alignment requirement; S is not a complete world-frame state oracle. \\
\addlinespace[4pt]
LightWar &
V: launch-site preview and per-shot first-person frame stacks; launch coordinates, trajectory law, action ranges and tolerance. Target coordinates and numerical miss distance are withheld. &
S: target apex coordinates, ground height at the target and shot telemetry (landing, miss vector and sampled ground heights), without images. C adds the visual observations. V+Telemetry adds scalar miss distance to V. &
Target localization and use of trajectory/terrain feedback. A global terrain height map is not supplied. Instruction variants are separate configurations. \\
\addlinespace[4pt]
Billiards &
V: settled table image, shot API, coordinate conventions, physical-behaviour guidance and pocket/scratch counts. &
S: numerical table/rail/pocket parameters, ball properties and current ball positions, without images. C adds the settled image. Position-query and aim-line-preview harnesses are separate variants. &
Use of explicit ball geometry and physical parameters. Preview or query variants also alter feedback or interaction. \\
\addlinespace[4pt]
MazeNavFPS &
V: current first-person view and previous action's blocked/ok feedback, without coordinates, raycasts or a harness-maintained position log. &
VP: image, current position/facing and explored-position log. RP: raycast distances plus position/log, without images. VRP: image plus RP. Goal coordinates and a full maze map are withheld. &
Local visual navigation, spatial memory support and range sensing. RP is not simply VP with its image removed; batch/single remain separate protocols. \\
\addlinespace[4pt]
TunnelNav3D &
The released aim-and-fly variant uses images and blocked/ok feedback. Its VP variant supplies images, position/facing and explored-position history. &
The reported RP and VRP configurations denote raycasts plus position history, without and with images, respectively. The current release exposes VP, aim-and-fly and the equal-move VP/VRP protocols. &
Navigation with different sensing, memory and control support. The channel name must be read together with the control protocol. \\
\addlinespace[4pt]
Scene\-Grounding &
Target reference image and current view, current camera pose, action definitions and movement/blocked feedback. The target camera pose is withheld. &
The released view-recovery task uses visual observations throughout; no corresponding image-free S condition is defined. &
Recovering the reference viewpoint using visual landmarks with known current pose, not recovering all pose information from pixels. \\
\addlinespace[4pt]
Object\-Grounding &
Both released V/C-labelled variants supply the target description, camera pose, screen-to-world axis mapping and marker coordinates. The grid family also supplies an anchor index/colour/coordinate/distance table. Target coordinates are withheld. &
Both use images. The \texttt{bboxv2} variant changes only the rendered labels (size, position and backing); it does not newly introduce the coordinate table. No image-free S condition is defined. &
Visual target identification and marker-guided localization under different label presentations. This release contrast is not the addition of a previously absent coordinate channel. \\
\addlinespace[4pt]
Curvy\-Billiards &
Released rolling-ball V: surface/target images, grid geometry, ball radius, friction, fixed start and speed bound; action is launch angle and speed. &
S: numerical surface parameters, target geometry and rollout log, without images. C includes both. Reference solutions are omitted from model input. &
Surface/target interpretation versus using explicit geometry to choose a launch. This row specifies the released rolling-ball entry point. \\
\addlinespace[4pt]
DropTo\-Target &
Released surface-design V: scene images, grid geometry, ball radius/start and initial surface height. The model supplies a surface-height function. &
S: the same setup plus target coordinates/radius and rollout log, without images. C includes both. All conditions receive verifier verdicts and action-format errors. &
Target localization and surface design for the ball's first ground contact. This is not cloth-parameter recovery. \\
\end{longtable}
\addtocounter{table}{-1}% This unnumbered inventory must not renumber existing result tables.
\endgroup

\paragraph{Pixel-only control for PlaceOnBottom.} To separate the supplied geometry from visual identification, we ran \textit{PlaceOnBottom} on 50 seeds with images and anonymous object identifiers only (\href{https://github.com/VITA-Group/VeriWorld/blob/paper-artifacts-2026-11-06-rc3/veriworld/benchmark/interactive_control/place_on_bottom/pure_visual.py}{released harness}). In V, every round lists the current world position and size of each object; in this control, no positions, sizes, camera pose or coordinate reference are given at any round, and the only feedback is the next screenshot and an echo of the model's own last command. The action interface (\texttt{probe} and \texttt{move\_to} with absolute world coordinates), the round budget (8) and the verifier are unchanged. Pass@1 is 0.06 for Gemini-3.1-Pro ($n=50$) and 0.00 for Opus-4.5 ($n=48$; two episodes were unscorable because an object left the scene after repeated probes). On the seeds also run under V, the values are 0.10 vs.\ 0.85 (Gemini-3.1-Pro, $n=20$) and 0.00 vs.\ 0.34 (Opus-4.5, $n=47$). Because \texttt{move\_to} requires absolute coordinates without any reference frame, this control mainly measures recovering a world frame from images rather than object identification alone.

\paragraph{Pixel-only controls across tasks.} Table~\ref{tab:pixel_only} lists all pixel-only controls. The \textit{LightWar} control (\href{https://github.com/VITA-Group/VeriWorld/blob/paper-artifacts-2026-11-06-rc3/veriworld/benchmark/trial_error_prediction/light_war/pixel_only.py}{released harness}) removes the launch-site coordinates, the trajectory law and gravity, the tolerance, the beacon range and tower height, and V's numeric aiming defaults, keeping the images, action names and units, the compass convention and the speed range. The \textit{BoxFold} control (\href{https://github.com/VITA-Group/VeriWorld/blob/paper-artifacts-2026-11-06-rc3/veriworld/benchmark/trial_error_prediction/box_fold/pure_visual.py}{released harness}) removes the layout name, the flat-layout and hinge tables, the edge-to-axis rules, the cube size, the target centre and the tolerance, drops the native fold echo, and lists only the allowed hinge identifiers; because the render does not label faces or hinges, the model can learn what a hinge does only by folding.

\begin{table}[h]
\centering
\small
\caption{\textbf{Pixel-only controls.} Pass@1 with images and anonymous identifiers only; no coordinates, layout or hinge tables, or numeric task parameters are given.}
\label{tab:pixel_only}
\begin{tabular}{@{}lllrr@{}}
\toprule
\textbf{Task} & \textbf{Action interface} & \textbf{Model} & \textbf{Pass@1} & $n$ \\
\midrule
PlaceOnBottom & absolute world coordinates & Gemini-3.1-Pro & 0.06 & 50 \\
 & & Opus-4.5 & 0.00 & 48 \\
\addlinespace
BoxFold & hinge identifiers & Gemini-3.1-Pro & 0.00 & 20 \\
 & & Sonnet-4.6 & 0.00 & 42 \\
\addlinespace
LightWar & launcher-centred launch & Gemini-3.1-Pro & 0.63 & 30 \\
 & & Sonnet-4.6 & 0.00 & 28 \\
 & & GPT-5.2 & 0.00 & 9 \\
\bottomrule
\end{tabular}
\end{table}

\paragraph{Reading condition differences.}
These specifications make the supplied and withheld information explicit. Prompt instructions and feedback can also differ between task-specific conditions; S$-$V therefore measures performance under the stated configurations rather than an isolated perceptual error. Likewise, C$-$S does not by itself identify a unique multimodal-fusion mechanism.

\section{Specialized Results}
\label{app:specialized_results}

Some task families are central to \ours but should not be collapsed into the core V/S/C table in the main text.
Photoreal scene grounding is visual-only by construction, and pose-assisted navigation uses auxiliary information channels.
We report them separately for semantic clarity: these rows exercise the photoreal grounding and navigation-memory regimes that motivate the benchmark infrastructure.

\subsection{Specialized Tracks: Photoreal Grounding and Pose-Assisted Navigation}

\begin{table}[!ht]
\centering
\caption{
\textbf{Specialized and extension-track results.}
These rows preserve task families whose semantics differ from the clean V/S/C core table in the main text.
Photoreal grounding is visual-only by construction.
\textit{TunnelNav3D} uses auxiliary navigation variants: VP = rendered observation + position history, RP = raycast + position without rendered observation, and VRP = rendered observation + raycast and position.
}
\label{tab:specialized_results}
\small
\setlength{\tabcolsep}{2.5pt}
\begin{tabular}{@{}lllcccc@{}}
\toprule
\textbf{Track} & \textbf{Task} & \textbf{Model} & \textbf{Cond.~A} & \textbf{Cond.~B} & \textbf{Cond.~C/Aux} & \textbf{Note} \\
\midrule
\multicolumn{7}{@{}l}{\textit{Photoreal Grounding: visual-only by construction}} \\
\midrule
\multirow{6}{*}{Photo} & \multirow{6}{*}{SceneGrounding}
& Gemini-3.1-Pro & Step/turn=0.60 & Walk/view=0.00 & -- & V only \\
& & Gemini-2.5-Flash & Step/turn=0.70 & Walk/view=0.00 & -- & V only \\
& & Sonnet-4.6 & Step/turn=0.60 & Walk/view=0.00 & -- & V only \\
& & Opus-4.5 & Step/turn=0.50 & Walk/view=0.00 & -- & V only \\
& & Grok-4.3 & Step/turn=0.30 & Walk/view=0.20 & -- & V only \\
& & GPT-4.1 & Step/turn=0.20 & Walk/view=0.10 & -- & V only \\
\midrule
\multicolumn{7}{@{}l}{\textit{Pose-assisted Navigation: auxiliary variants}} \\
\midrule
\multirow{3}{*}{Tunnel} & \multirow{3}{*}{TunnelNav3D}
& Gemini-3-Pro & VP=0.75 & RP=0.00 & VRP=0.75 & RP$-$VP=-0.75 \\
& & GPT-4.1     & VP=0.25 & RP=0.00 & VRP=0.62 & RP$-$VP=-0.25 \\
& & Claude Opus & VP=0.25 & RP=0.00 & VRP=0.25 & RP$-$VP=-0.25 \\
\bottomrule
\end{tabular}
\vspace{2pt}
\begin{minipage}{0.96\linewidth}
\footnotesize
\textit{SceneGrounding}: $n=10$ per cell (two interior levels $\times$ seeds 0--4); Step/turn and Walk/view are two collision-aware action interfaces of the same view-recovery task: Step/turn moves forward or backward by a chosen distance and turns by relative angles, while Walk/view walks forward and sets an absolute view direction. Only the Step/turn verifier also checks pitch.
For photoreal grounding, no structured-only condition is defined because the rendered screenshot is the task signal.
For TunnelNav3D, VP/RP/VRP are auxiliary variants and should not be read as core V/S/C conditions.
\end{minipage}
\end{table}

\begin{table}[!ht]
\centering
\caption{
\textbf{MazeNavFPS auxiliary information-channel sweep, 5$\times$5 grids, 9-seed extension.}
Maze navigation does not cleanly fit the core V/S/C taxonomy: a closed-loop ReAct agent inevitably accumulates an action-history memory across rounds, so even the rendered-observation-only condition retains an \emph{implicit} structured channel.
We sweep four auxiliary information channels (all under batched action control, $Bf$), holding the model lineup fixed at four frontier VLMs:
\textbf{V} -- rendered observation only; no harness-tracked position log;
\textbf{RP} -- 4-direction raycast wall distances $+$ harness-tracked position log; no rendered observation;
\textbf{VP} -- rendered observation $+$ position log;
\textbf{VRP} -- rendered observation $+$ raycast $+$ position log.
We restrict to the hardest 5$\times$5 grid for an apples-to-apples shape comparison and use nine seeds $\{0,1,2,3,4,5,6,7,42\}$ that produce different maze topologies.
Each per-model cell reports Pass@1 over valid episodes; \textbf{Avg} pools them across the four models. Each cell has $n=9$ except RP (Gemini 7, Opus 4, GPT-5.2 5), VP (Opus 7) and VRP (Opus 4, GPT-5.2 8).
}
\label{tab:maze_sweep}
\small
\setlength{\tabcolsep}{4pt}
\begin{tabular}{@{}lccccc@{}}
\toprule
\textbf{Condition} & \textbf{Gemini-3.1-Pro} & \textbf{Sonnet} & \textbf{Opus} & \textbf{GPT-5.2} & \textbf{Avg} \\
\midrule
V \quad (rendered only)                       & 0.56 & 0.78 & 0.56 & 0.33 & 0.56 \\
RP (raycast + position log)       & 0.86 & 0.44 & 0.00 & 0.40 & 0.48 \\
VP \quad (rendered + position log)            & 0.44 & 0.78 & 0.43 & 0.67 & 0.59 \\
VRP \quad (rendered + raycast + position log) & 1.00 & 0.67 & 0.25 & 0.75 & 0.73 \\
\bottomrule
\end{tabular}
\vspace{2pt}
\begin{minipage}{0.96\linewidth}
\footnotesize
\textbf{Gemini-3.1-Pro under VRP is the headline}: 1.00 Pass@1 across nine distinct maze topologies, the only condition--model cell at 1.00 in the sweep, indicating that combining the rendered scene with raycast and position log saturates the harness for the strongest model.
\textbf{V has an aggregate pass rate of 0.56}: the evaluated mazes include topologies where pure-vision agents lose track once the rendered cell is small.
\textbf{RP has an aggregate pass rate of 0.48}: Gemini-3.1-Pro reaches 0.86 from raycast alone, consistent with raycast being a strong cue for this model on novel shapes.
\textbf{VP and VRP do not uniformly dominate}: Sonnet and Opus do not improve over V when extra channels are added, whereas GPT-5.2 improves under both and Gemini-3.1-Pro only under VRP.
A per-model breakdown including additional models is provided in App.~\ref{app:maze_results_section}.
\end{minipage}
\end{table}

Table~\ref{tab:specialized_results} reports results for photoreal scene grounding and pose-assisted navigation.
The photoreal grounding tasks have no structured-only condition: the goal is defined by semantic landmarks in a rendered 3D scene, such as recovering a camera pose in an interior level or localizing a named object in a kitchen, while the visual input still includes interface metadata such as the current camera pose (Appendix~\ref{app:task_inputs}). In \textit{SceneGrounding}, the Step/turn interface succeeds more often than Walk/view for all six models, even though both move with collision and Step/turn is also checked on pitch; because the two interfaces differ in both their commands and their verifier, the comparison does not isolate a single factor.
The navigation rows and the MazeNavFPS sweep in Table~\ref{tab:maze_sweep} show why auxiliary notation is necessary: these tasks expose position history or other navigation aids, so they should not be labeled as pure V even when rendered observations are present.

\section{Additional Harness Diagnostics}
\label{app:harness_extra}

Table~\ref{tab:case_maze} in the main text compares batch and single-step control under an equal primitive-move budget. Batch and single share a budget fixed before the runs at the 90th percentile of moves used by successful batch episodes under the 50-call budget (125, 170 and 210 moves for $3{\times}3$, $4{\times}4$ and $5{\times}5$ grids). Every command counts against the budget, a final batch is truncated to the remaining moves, and the goal is checked after every move; single uses the standard single-step harness with its call cap equal to the budget (see the released \href{https://github.com/VITA-Group/VeriWorld/tree/paper-artifacts-2026-11-06-rc3/veriworld/benchmark/interactive_control/mazenavfps/harness_structured/vp_bf_eq}{VP} and \href{https://github.com/VITA-Group/VeriWorld/tree/paper-artifacts-2026-11-06-rc3/veriworld/benchmark/interactive_control/mazenavfps/harness_structured/vrp_bf_eq}{VRP} equal-move harnesses). The two protocols still differ in feedback schedule, because batch observes the scene once per batch. All 240 cells of each variant are scored. Per model, the exact McNemar $p$ is 0.003, 0.22, 0.04 and 0.34 under VP and 0.0005, 0.001, 0.125 and 0.002 under VRP (Gemini-3.1-Pro, Sonnet-4.6, GPT-5.2, GPT-4.1). Table~\ref{tab:maze_equal_budget} gives per-grid results: batch leads at every grid size under both variants, most clearly on $4{\times}4$ and $5{\times}5$ grids. Batch used fewer primitive moves than single and far fewer model calls (18.57 vs.\ 120.32 under VP; 20.48 vs.\ 105.87 under VRP). Table~\ref{tab:maze_call_budget} reports the comparison under an equal budget of 50 model calls, in which batch can execute more primitive moves than single; batch exceeds single for every reported model and variant there as well.

\paragraph{Response parsing.} In 17 of 30 GPT-4.1 VRP single-step episodes, the model emitted a turn action without the required command field (e.g., \verb|{"move": {"turn": 180}}|); the single-step harness cannot execute it and ends the episode, which Table~\ref{tab:case_maze} scores as a failure. Re-running both GPT-4.1 VRP protocols with a parser that maps such turn, forward and backward forms to the command schema gives 0.77 (batch) vs.\ 0.67 (single), with 4 vs.\ 1 discordant mazes ($p=0.375$). GPT-4.1 under VRP therefore does not provide evidence for batch over single; without GPT-4.1, pooled VRP is 0.77 vs.\ 0.46 (29 vs.\ 1, $p<0.001$). Apart from one GPT-4.1 VP single-step episode, no other model or protocol ended an episode this way. Restricting the comparisons to episode pairs in which neither episode had a parsing issue leaves the pooled contrasts unchanged (VP 29 vs.\ 5, VRP 29 vs.\ 1, both $p<0.001$; tunnel paired differences identical).

\begin{table}[h]
\centering
\small
\caption{\textbf{Equal primitive-move budget on grid mazes, by grid size.} Pass@1 pooled over four models (Gemini-3.1-Pro, Sonnet-4.6, GPT-5.2, GPT-4.1) and 10 seeds per grid ($n=40$ per cell; $n=120$ for All), with mean primitive moves in parentheses. Discordant pairs count matched mazes passed by only one protocol; $p$ is the exact McNemar test.}
\label{tab:maze_equal_budget}
\begin{tabular}{@{}lcccc@{}}
\toprule
& \multicolumn{2}{c}{\textbf{VP}} & \multicolumn{2}{c}{\textbf{VRP}} \\
\cmidrule(lr){2-3}\cmidrule(lr){4-5}
\textbf{Grid} & \textbf{Batch} & \textbf{Single} & \textbf{Batch} & \textbf{Single} \\
\midrule
$3{\times}3$ & 0.92 (42.20) & 0.85 (51.20) & 0.90 (37.65) & 0.72 (43.08) \\
$4{\times}4$ & 0.75 (93.85) & 0.48 (125.90) & 0.82 (94.02) & 0.42 (104.98) \\
$5{\times}5$ & 0.50 (167.35) & 0.20 (183.82) & 0.50 (155.50) & 0.12 (169.12) \\
\midrule
All & 0.72 (101.13) & 0.51 (120.31) & 0.74 (95.72) & 0.42 (105.72) \\
Batch-only vs.\ single-only & \multicolumn{2}{c}{32 vs.\ 6, $p<0.001$} & \multicolumn{2}{c}{39 vs.\ 1, $p<0.001$} \\
\bottomrule
\end{tabular}
\end{table}

\begin{table}[h]
\centering
\caption{
\textbf{Grid mazes under an equal budget of 50 model calls.} Each cell shows Pass@1, with average rounds (model calls) in parentheses. The batch prompt asks for 3--6 commands per call and single executes one, so batch can execute more primitive moves. VP = rendered observation + position history; VRP = VP + a 4-direction raycast scan.
}
\label{tab:maze_call_budget}
\small
\setlength{\tabcolsep}{3pt}
\begin{tabular}{@{}lccccc@{}}
\toprule
& \multicolumn{2}{c}{\textbf{VP}} & \multicolumn{2}{c}{\textbf{VRP}} & \\
\cmidrule(lr){2-3}\cmidrule(lr){4-5}
\textbf{Model} & \textbf{Batch} & \textbf{Single} & \textbf{Batch} & \textbf{Single} & \textbf{N} \\
\midrule
Gemini 3 Pro & 0.89 (21.44) & 0.44 (35.78) & 0.89 (29.78) & 0.67 (33.17) & 9 \\
Sonnet       & 0.89 (20.89) & 0.22 (46.11) & 0.67 (27.33) & 0.33 (44.00) & 9 \\
GPT-5.2      & 0.89 (14.56) & 0.33 (40.89) & 0.78 (37.11) & 0.11 (49.22) & 9 \\
Claude Opus & -- & -- & 0.44 (41.33) & 0.00 (50.00) & 9 \\
GPT-4.1      & 0.67 (36.67) & 0.22 (43.89) & 0.78 (25.78) & 0.50 (38.25) & 9 \\
Qwen3-235B   & 0.44 (39.78) & 0.22 (44.11) & 0.33 (35.67) & 0.00 (50.00) & 9 \\
Doubao       & 0.22 (47.22) & 0.00 (50.00) & 0.11 (49.00) & 0.00 (50.00) & 9 \\
\bottomrule
\end{tabular}
\vspace{2pt}
\begin{minipage}{0.96\linewidth}
\footnotesize
N is the number of mazes (grids 3, 4, 5 $\times$ seeds 0, 7, 42); each cell uses the earliest valid run per maze, and average rounds are mean model calls over those runs. API or infrastructure failures are excluded: effective $N=6$ for Gemini 3 Pro VRP Single and $N=8$ for the VRP Single cells of Claude Opus, GPT-4.1, Qwen3-235B and Doubao.
\end{minipage}
\end{table}
Two complementary harness studies extend this finding from different angles: on \textit{TunnelNav3D}, the action protocol (active aim-and-fly vs.\ batched plans vs.\ single steps) changes success rates substantially under an equal primitive-move budget; on \textit{Billiards}, adding visual self-feedback (a rendered aim-line) flips outcome without revealing any new hidden scene state.

\subsection{Curved Tunnels: The Action Protocol Strongly Affects Success}
\label{app:harness_tunnel}

\begin{table}[!b]
\centering
\caption{
\textbf{Curved tunnels under an equal budget of 80 primitive moves: active control is best or tied for best for every model.} Pass@1 with mean primitive moves in parentheses. All three protocols share a budget of 80 primitive engine commands; the Active, Batch and Single harnesses are otherwise unchanged. Each cell pools 8 tunnel configurations (2 radii $\times$ 2 textures $\times$ 2 cardinal settings) $\times$ 2 repeats ($n=16$; $n=14$ for Claude Opus-4.5 VRP Batch); repeats are not independent seeds.
VP = rendered observation + position history; VRP = VP + raycast.
Active means orient-then-move each turn; Batch means multi-move plan per turn; Single means one free move per turn.
}
\label{tab:tunnel_equal_budget}
\small
\setlength{\tabcolsep}{2.5pt}
\begin{tabular}{@{}lcccccc@{}}
\toprule
& \multicolumn{3}{c}{\textbf{VP}} & \multicolumn{3}{c}{\textbf{VRP}} \\
\cmidrule(lr){2-4}\cmidrule(lr){5-7}
\textbf{Model} & \textbf{Active} & \textbf{Batch} & \textbf{Single} & \textbf{Active} & \textbf{Batch} & \textbf{Single} \\
\midrule
Gemini-3.1-Pro & 0.88 (41.56) & 0.06 (77.88) & 0.12 (77.50) & 0.88 (44.25) & 0.50 (54.69) & 0.44 (67.69) \\
Claude Opus-4.5 & 0.81 (43.88) & 0.31 (67.81) & 0.19 (68.00) & 0.69 (55.00) & 0.14 (75.21) & 0.69 (52.56) \\
GPT-4.1 & 0.81 (37.94) & 0.00 (80.00) & 0.19 (70.88) & 0.56 (58.44) & 0.06 (77.81) & 0.12 (71.50) \\
\midrule
All & 0.83 (41.12) & 0.12 (75.23) & 0.17 (72.12) & 0.71 (52.56) & 0.24 (68.98) & 0.42 (63.92) \\
\bottomrule
\end{tabular}
\end{table}

Table~\ref{tab:tunnel_equal_budget} compares the three protocols with an equal budget of 80 primitive moves (see the \href{https://github.com/VITA-Group/VeriWorld/tree/paper-artifacts-2026-11-06-rc3/veriworld/benchmark/interactive_control/tunnel_nav3d/harness_structured/legacy_eq}{released equal-move harness}). Active control is best or tied for best for every model under both VP and VRP (pooled: active 0.77, batch 0.18, single 0.29; paired difference active$-$batch $+0.60$, 95\% CI $[0.48, 0.71]$; active$-$single $+0.48$ $[0.35, 0.60]$; cluster bootstrap over model $\times$ variant $\times$ configuration), while using fewer primitive moves (46.84 on average, vs.\ 72.17 for batch and 68.02 for single).

\subsection{Billiards: Self-Feedback Is a Harness Variable, Not New Hidden Information}
\label{app:harness_aimline}

\begin{table}[!ht]
\centering
\caption{
\textbf{Billiards: visual self-feedback improves performance without revealing hidden scene state.}
V+REPL renders the agent's proposed shot as a 1500\,cm aim-line and allows up to three revisions before firing.
The action vocabulary and hidden world state are unchanged; the harness adds self-visualization of the proposed action.
Pass@1 is the mean fraction pocketed on the hard preset (10 balls $\times$ 10 rounds, 3-seed mean). Bold $\Delta \geq +0.20$.
}
\label{tab:case_aimline}
\small
\begin{tabular}{@{}lrrr@{}}
\toprule
\textbf{Model} & \textbf{V} & \textbf{V+REPL} & \textbf{$\Delta$} \\
\midrule
Flash        & 0.40 & 0.80 & \textbf{+0.40} \\
Sonnet       & 0.33 & 0.67 & \textbf{+0.33} \\
Gemini-3-Pro & 0.63 & 0.80 & +0.17 \\
Grok         & 0.10 & 0.27 & +0.17 \\
Doubao       & 0.23 & 0.37 & +0.13 \\
GPT-4o       & 0.47 & 0.57 & +0.10 \\
GPT-5.2      & 0.20 & 0.27 & +0.07 \\
Qwen3-235B   & 0.10 & 0.17 & +0.07 \\
Moonshot     & 0.10 & 0.17 & +0.07 \\
Qwen-Max     & 0.07 & 0.00 & -0.07 \\
GPT-4.1      & 0.33 & 0.20 & -0.13 \\
\bottomrule
\end{tabular}
\vspace{2pt}
\begin{minipage}{0.96\linewidth}
\footnotesize
Rows are ordered by descending $\Delta$. $N=30$ balls per cell (10 balls $\times$ 3 seeds).
\end{minipage}
\end{table}

Table~\ref{tab:case_aimline} isolates a different protocol effect.
V+REPL lets the agent see a rendered aim-line for its own proposed shot before committing, but does not reveal additional hidden scene state.
This simple self-feedback raises performance for 9 of 11 models, including large gains for Flash (0.40$\to$0.80) and Sonnet (0.33$\to$0.67), while GPT-4.1 regresses.
The result shows that a harness can improve or degrade performance in a model-dependent way even when the task and hidden state are unchanged.

\subsection{LightWar: Telemetry vs.\ Vision-Only Input Channels}
\label{app:harness_lightwar}

\begin{figure}[!ht]
\centering
\includegraphics[width=0.92\linewidth]{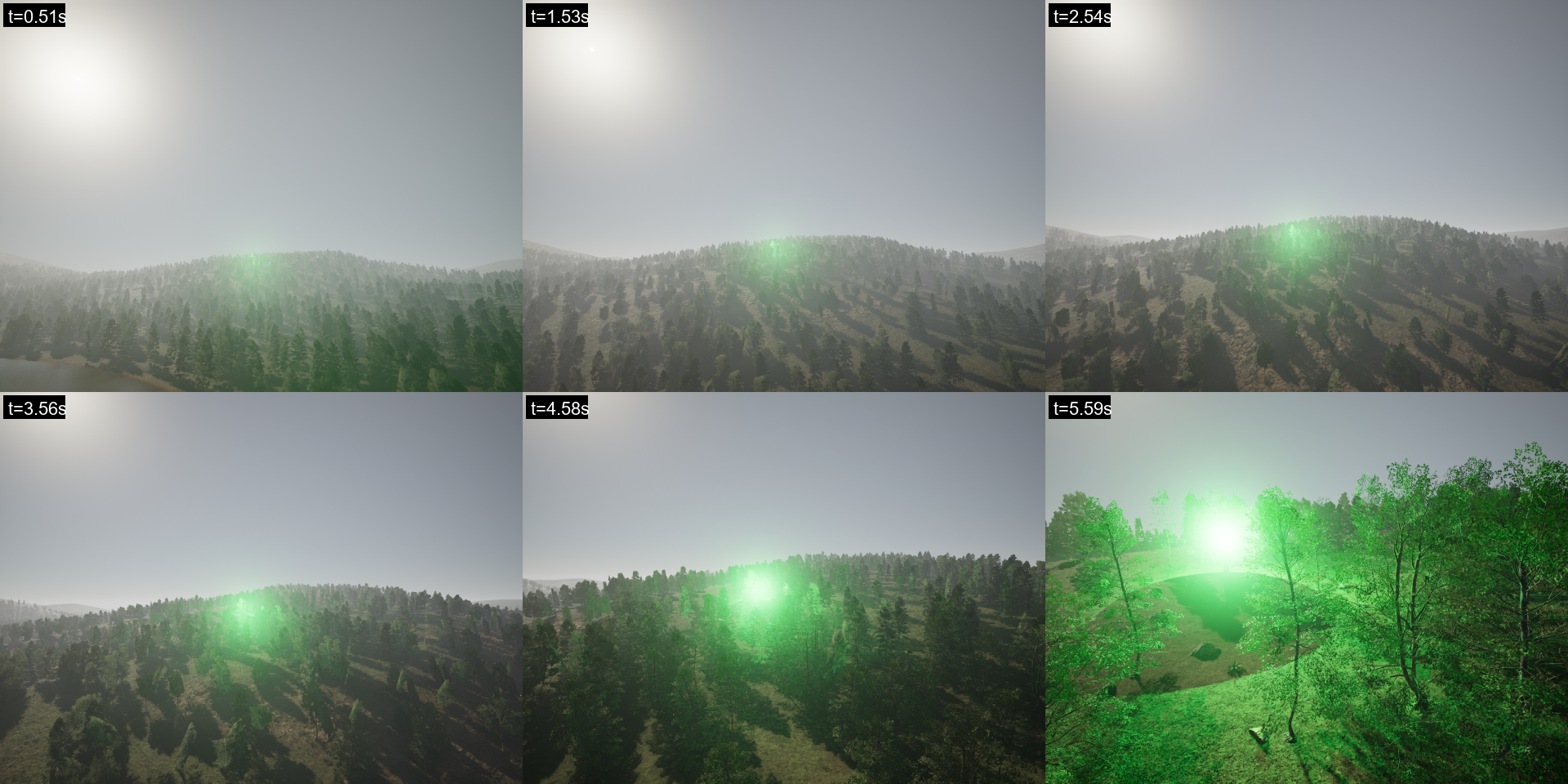}
\caption{
\textbf{LightWar successful arc, V$+$Telemetry condition.}
A 2$\times$3 stack of arc replay frames sampled at $t=0.51, 1.53, 2.54, 3.56, 4.58, 5.59$\,s after launch, taken from a Gemini~3.1~Pro Pass@1 episode on seed=0 (\texttt{hit\_forest\_hilltop\_beacon}).
The agent fires a glowing projectile from a low-elevation forest clearing (frame~1) and tracks the ballistic arc until it intersects a hilltop beacon (frame~6, bright impact flare).
Each frame is the harness's per-tick render passed back to the agent at the next round; together they form the visual evidence the agent uses to plan the next shot.
Per-round screenshots like these are part of the V condition; V$+$Telemetry additionally reports the scalar miss distance of each shot.
Without the miss distance (V-only), the mid-tier panel models (Sonnet, Opus, Kimi, Qwen3-VL-30B) drop from 0.33 to 0.00 Pass@1.
}
\label{fig:lightwar_arc_pass}
\end{figure}

The LightWar results in Table~\ref{tab:core_results} (main text) compare the V/S/C conditions.
The matched results use 30 seeds for Gemini-3.1-Pro, 25 for Sonnet-4.6, and 9 for GPT-5.2, totaling 192 paired episodes.
The primary per-condition analysis includes all valid episodes and rounds to the same two-decimal scores as the matched comparison in Table~\ref{tab:core_results}.
Its V, S, and C sample sizes are respectively $(30,30,30)$ for Gemini, $(28,30,27)$ for Sonnet, and $(9,10,10)$ for GPT-5.2; missing runs are not scored as failures.
Independently, this appendix reports a \emph{separate harness ablation} that compares two narrower input-channel configurations on the same task with 3 seeds per cell. This ablation isolates the input-channel toggle while holding task, model, and round budget fixed. Numbers in this appendix do \emph{not} substitute for the V/S/C entries in Table~\ref{tab:core_results}; they instead extend the harness-as-confound analysis to a third axis (input-channel set), complementing action span (\textit{MazeNavFPS}, Sec.~\ref{sec:case_studies}) and intra-round feedback (\textit{Billiards}, App.~\ref{app:harness_aimline}).

Figure~\ref{fig:lightwar_arc_pass} shows the per-tick frame stack for one passing episode in the V$+$Telemetry condition --- the kind of visual evidence the agent works from each round.

\begin{table}[t]
\centering
\caption{
\textbf{LightWar (outdoor projectile-aim): reporting the miss distance alongside vision changes outcomes for mid-tier models, but does not rescue weak ones.}
Same task instance, same model, same maximum-rounds budget; only whether the miss distance is reported changes.
V$+$Telemetry additionally reports the scalar miss distance of each shot; V-only is the V condition of Appendix~\ref{app:task_inputs}, which withholds it.
Pass@1 over 3 seeds $\{0, 1, 2\}$.
}
\label{tab:case_lightwar}
\small
\setlength{\tabcolsep}{6pt}
\begin{tabular}{@{}lccc@{}}
\toprule
\textbf{Model} & \textbf{V$+$Telemetry} & \textbf{V-only} & \textbf{$\Delta$} \\
\midrule
Gemini~3.1~Pro & 0.67 & 0.67 & \phantom{$+$}0.00 \\
Sonnet         & 0.33 & 0.00 & -0.33 \\
GPT-5.2        & 0.00 & 0.00 & \phantom{$+$}0.00 \\
\bottomrule
\end{tabular}
\vspace{2pt}
\begin{minipage}{0.96\linewidth}
\footnotesize
$\Delta$ = V-only $-$ V$+$Telemetry. Three representative models from the harness-comparison panel. Across the full panel (35 episodes per condition), V$+$Telemetry achieves 0.17 and V-only 0.06; the ratio is dominated by mid-tier models like Sonnet, Opus, Kimi, and Qwen3-VL-30B that pass 0.33 with telemetry and drop to 0.00 without.
\end{minipage}
\end{table}

The harness toggle controls only one additional signal: V$+$Telemetry reports the scalar miss distance of each shot, and V-only withholds it.
At this 3-seed scale Table~\ref{tab:case_lightwar} exposes a non-monotone three-band response across model strength.
\textit{Top of the panel:} Gemini~3.1~Pro achieves 0.67 in both conditions; the V-only screenshots already encode enough projectile geometry for it to plan, so removing telemetry does not hurt.
\textit{Middle of the panel:} Sonnet drops from 0.33 to 0.00 once telemetry is removed --- it can use the reported miss distance but cannot recover the same information from the visual stream at $N{=}3$.
\textit{Bottom of the panel:} GPT-5.2 achieves 0.00 in both conditions. In the core V/S/C comparison (Table~\ref{tab:core_results}), it achieves 0.00 under V and 1.00 under S and C.
Across the full 12-model panel at this scale (35 episodes per condition), V$+$Telemetry achieves \textbf{0.17} Pass@1 and V-only achieves \textbf{0.06}; the 0.11 gap is explained almost entirely by four mid-tier models (Sonnet, Opus, Kimi, Qwen3-VL-30B) that pass once per three seeds with telemetry and never without it.
This three-band response (vision-robust top tier, harness-sensitive middle band, harness-insensitive bottom tier at small $N$) is consistent with the maze and tunnel results in Sec.~\ref{sec:case_studies} and Apps.~\ref{app:harness_tunnel},~\ref{app:harness_aimline}: the harness is a first-class confound, but its leverage is highest for models that have the reasoning capacity yet lack the perceptual robustness to recover the same information from pixels.

\subsection{ObjectGrounding: Reporting Tracks and Full Harness Panels}
\label{app:harness_object_ground}

\paragraph{How this task fits the reporting-tracks taxonomy.}
\textit{ObjectGrounding} sits in the \emph{photoreal grounding specialized track}: the rendered scene is itself the task signal, so a structured-only condition (S) is not defined---semantic targets such as ``the wooden spoon'' versus ``the black spoon'' cannot be disambiguated without the rendered scene.
The remaining V/C contrast in the main table (Table~\ref{tab:core_results}, grid-bbox harness pair) is therefore the \emph{core information-condition track} restricted to two clean conditions.
A second harness pair (single-bbox), reported below as a contrast, runs the same V/C ablation under a sparser action-interface scaffold and so falls into the \emph{harness-diagnostic track}: it isolates whether the symbolic-index channel's helpfulness is itself harness-dependent.

\begin{figure}[h]
\centering
\setlength{\tabcolsep}{2pt}
\begin{adjustbox}{max width=\linewidth}
\begin{tabular}{cccc}
\multicolumn{2}{c}{\textbf{Single-marker family}} & \multicolumn{2}{c}{\textbf{Grid-marker family (9 anchors)}} \\
\addlinespace[2pt]
\includegraphics[width=0.22\linewidth]{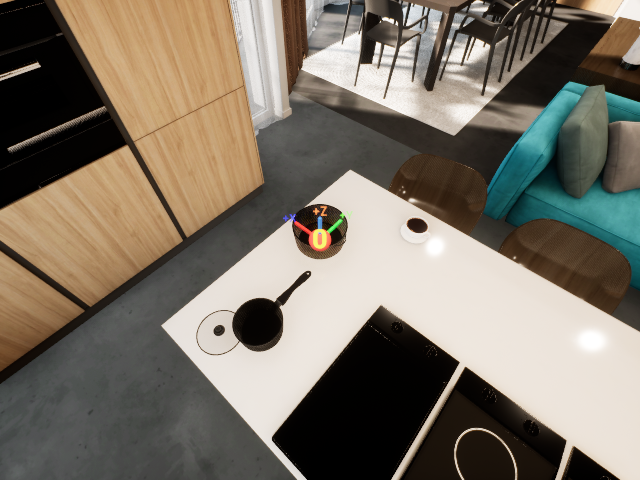} &
\includegraphics[width=0.22\linewidth]{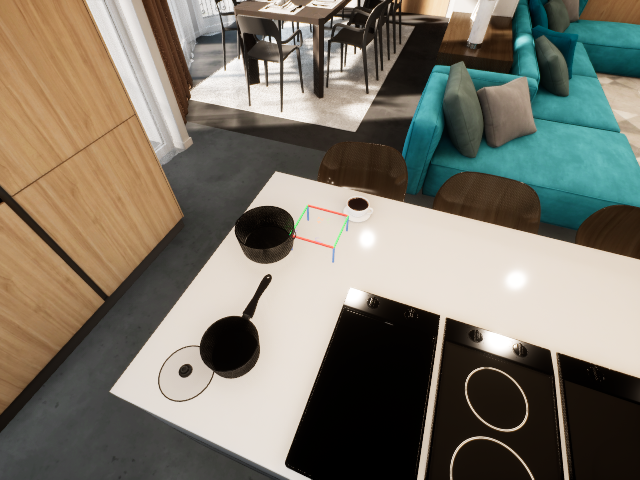} &
\includegraphics[width=0.22\linewidth]{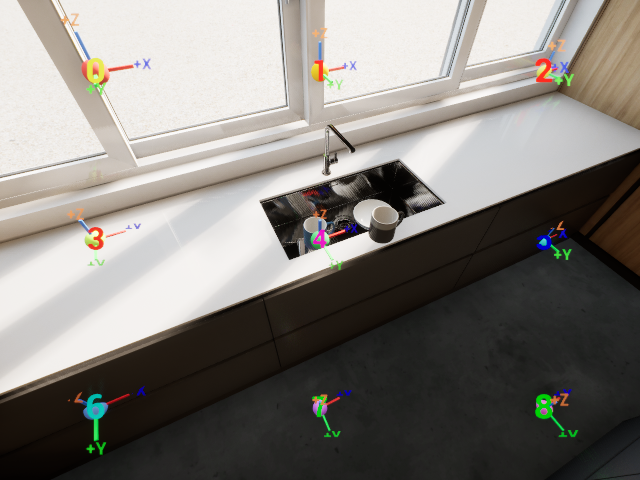} &
\includegraphics[width=0.22\linewidth]{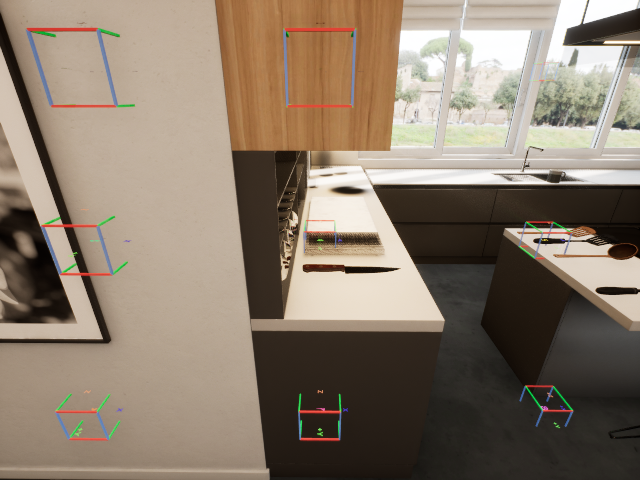} \\
\footnotesize\texttt{single\_iterate} & \footnotesize\texttt{single\_iterate\_offc.} & \footnotesize\texttt{grid\_iterate} & \footnotesize\texttt{grid\_iterate\_offc.} \\
\footnotesize ball + number & \footnotesize bbox \emph{no} number ($V$) & \footnotesize 9 balls + numbers & \footnotesize 9 bboxes \emph{no} numbers ($V$) \\
\addlinespace[6pt]
& \includegraphics[width=0.22\linewidth]{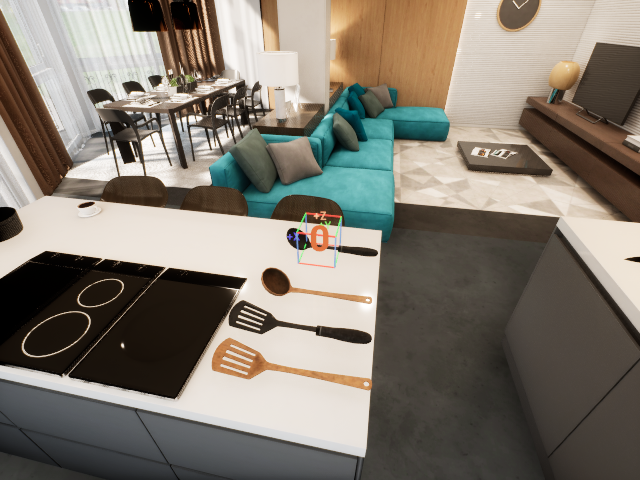} &
& \includegraphics[width=0.22\linewidth]{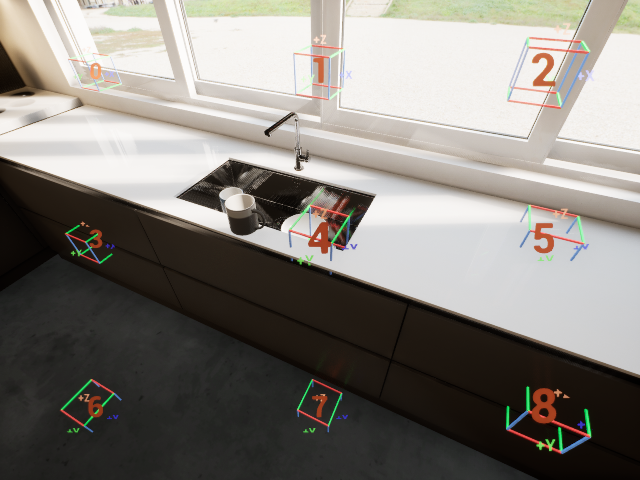} \\
& \footnotesize\texttt{single\_iterate\_offc.\_bboxv2} & & \footnotesize\texttt{grid\_iterate\_offc.\_bboxv2} \\
& \footnotesize bbox \emph{with} number ($C$) & & \footnotesize 9 bboxes \emph{with} numbers ($C$) \\
\end{tabular}
\end{adjustbox}
\caption{
\textbf{Six rendering variants of the \textit{ObjectGrounding} harness.}
The main-table V/C contrast (grid-marker family, fourth column) compares 9 bboxes with small index labels inside the boxes vs.\ the same bboxes with enlarged, camera-facing labels under an offcenter camera; the harness-diagnostic V/C contrast (single-marker family, second column) repeats the same label-rendering ablation under a sparser, single-bbox scaffold.
Auxiliary numbered-ball variants (first and third columns) are the original baselines from which the bbox variants were derived; they are not used for the V/C tables but are the scene context the seed sampler draws from.
Within each family, V and C differ only in how the index labels are rendered; both include the marker-to-world-coordinate table.
}
\label{fig:object_ground_harness_grid}
\end{figure}

\paragraph{Six harness rendering variants.}
Across both tracks, the underlying harness families are illustrated in Figure~\ref{fig:object_ground_harness_grid}.
The visualization choice (ball-with-axes vs.\ bbox; small or enlarged index labels) determines what symbolic channel is layered on top of the scene rendering, while the action-interface choice (single-marker vs.\ 9-anchor grid scaffold) determines how much pixel$\to$world structure is supplied before the grounding loop begins.

\subsubsection{Core V/C track (grid-marker family)}
\label{app:object_ground_core}

Table~\ref{tab:core_results} in the main text reports three frontier models under V (\texttt{harness\_grid\_iterate\_offcenter}: nine bboxes laid out in a $3\times3$ grid over the scene, each with a small index label inside the box, plus an index$\to$world-coordinate lookup table in the prompt) and C (\texttt{harness\_grid\_iterate\_offcenter\_bboxv2}: the same bboxes and table, with the labels enlarged, moved toward the camera and given a contrasting backing).
The full 12-model panel below extends that comparison.
Each cell is Pass@1 over $\sim$30 seeds drawn from a kitchen scene with 10 named objects, an offcenter camera that deliberately misses the target so the initial bboxes do not coincide with it, and an iterative grounding loop with up to 10 rounds.
The two harnesses share scene, target list, action interface, verifier, and per-seed seed list; the only signal that differs is how readable the index labels are.

\begin{table}[h]
\centering
\caption{
\textbf{ObjectGrounding core V/C panel (grid-marker family, 12 models).}
$C-V$ positive means making the nine candidate bboxes' index labels readable \emph{helps} performance: the agent can now reference a specific bbox by index instead of describing it visually.
9 of 12 models gain; macro-average $C-V = +0.10$.
The largest gains are on Claude Opus 4.5 ($+0.29$), Gemini-2.5-Flash ($+0.25$), GPT-5.2 ($+0.24$), and Gemini-3.1-Pro ($+0.21$); the visual scene already contains the answer, but most models cannot reliably describe \emph{which} of nine perceptually similar bboxes is closest until the symbolic index is supplied.
}
\label{tab:object_ground_full}
\small
\setlength{\tabcolsep}{4pt}
\begin{tabular}{@{}lrrr@{}}
\toprule
\textbf{Model} & \textbf{V (Pass@1, n)} & \textbf{C (Pass@1, n)} & \textbf{$C-V$} \\
\midrule
Gemini-3.1-Pro              & 0.69 ($n=36$) & 0.90 ($n=31$) & +0.21 \\
GPT-5-mini                  & 0.53 ($n=34$) & 0.47 ($n=30$) & -0.06 \\
GPT-5.2                     & 0.32 ($n=34$) & 0.57 ($n=30$) & +0.24 \\
Sonnet 4.6                  & 0.21 ($n=34$) & 0.27 ($n=30$) & +0.06 \\
GPT-4o                      & 0.26 ($n=34$) & 0.27 ($n=30$) & +0.00 \\
Claude Opus 4.5             & 0.15 ($n=34$) & 0.43 ($n=30$) & +0.29 \\
Kimi K2.5                   & 0.26 ($n=34$) & 0.33 ($n=30$) & +0.07 \\
Doubao Seed 1.6 Flash       & 0.03 ($n=33$) & 0.10 ($n=30$) & +0.07 \\
Qwen3-VL-30B-A3B            & 0.12 ($n=34$) & 0.23 ($n=30$) & +0.12 \\
Gemini-2.5-Flash            & 0.09 ($n=34$) & 0.33 ($n=30$) & +0.25 \\
Qwen3-Max                   & 0.12 ($n=33$) & 0.03 ($n=30$) & -0.09 \\
Grok-4                      & 0.00 ($n=33$) & 0.03 ($n=30$) & +0.03 \\
\midrule
Macro-average               & 0.23         & 0.33         & +0.10 \\
\bottomrule
\end{tabular}
\end{table}

The pattern is consistent with the rest of Table~\ref{tab:core_results}: when the index labels become readable, models that already possess the underlying spatial-reasoning competence get a clean lift, while those without do not.

\subsubsection{Harness-diagnostic track (single-marker family)}
\label{app:object_ground_harness}

Holding scene, target list, and verifier fixed, we now swap the action-interface scaffold from the 9-bbox grid down to a single bbox marker: \texttt{harness\_single\_iterate\_offcenter} for V (one bbox at the camera-axis hit, with a small index label inside it) and \texttt{harness\_single\_iterate\_offcenter\_bboxv2} for C (the same bbox with an enlarged, camera-facing label).
This pair differs from the core panel in exactly one variable: the agent has only one bbox to reason about per round, so a readable label has nothing to disambiguate from---there is no second candidate the index could distinguish against.

\begin{table}[h]
\centering
\caption{
\textbf{ObjectGrounding harness-diagnostic V/C panel (single-marker family, 12 models).}
Under the single-marker harness, $C-V$ flips sign: 9 of 12 models now lose performance when the lone bbox carries a readable label, with macro-average $C-V = -0.06$ (compared to $+0.10$ in Table~\ref{tab:object_ground_full}).
Without other bboxes to index against, a readable label is no longer a disambiguation channel---it is an extra symbol the agent must integrate into a still-unsolved visual localisation problem, and several models do worse rather than better.
}
\label{tab:object_ground_harness_grid}
\small
\setlength{\tabcolsep}{4pt}
\begin{tabular}{@{}lrrr@{}}
\toprule
\textbf{Model} & \textbf{V (Pass@1, n)} & \textbf{C (Pass@1, n)} & \textbf{$C-V$} \\
\midrule
Gemini-3.1-Pro              & 0.77 ($n=31$) & 0.68 ($n=34$) & -0.10 \\
GPT-5-mini                  & 0.63 ($n=30$) & 0.66 ($n=32$) & +0.02 \\
GPT-5.2                     & 0.63 ($n=30$) & 0.41 ($n=32$) & -0.23 \\
Sonnet 4.6                  & 0.47 ($n=30$) & 0.33 ($n=33$) & -0.13 \\
GPT-4o                      & 0.33 ($n=30$) & 0.21 ($n=33$) & -0.12 \\
Claude Opus 4.5             & 0.30 ($n=30$) & 0.39 ($n=33$) & +0.09 \\
Kimi K2.5                   & 0.30 ($n=30$) & 0.27 ($n=33$) & -0.03 \\
Doubao Seed 1.6 Flash       & 0.23 ($n=30$) & 0.12 ($n=32$) & -0.11 \\
Qwen3-VL-30B-A3B            & 0.23 ($n=30$) & 0.12 ($n=32$) & -0.11 \\
Gemini-2.5-Flash            & 0.20 ($n=30$) & 0.09 ($n=33$) & -0.11 \\
Qwen3-Max                   & 0.20 ($n=30$) & 0.28 ($n=32$) & +0.08 \\
Grok-4                      & 0.10 ($n=30$) & 0.09 ($n=32$) & -0.01 \\
\midrule
Macro-average               & 0.37         & 0.31         & -0.06 \\
\bottomrule
\end{tabular}
\end{table}

\paragraph{Cross-track summary: the same channel, opposite sign.}
Tables~\ref{tab:object_ground_full} and~\ref{tab:object_ground_harness_grid} report the same V/C ablation (numeric vs.\ no-numeric bbox labels) on the same scene with the same models, differing only in whether the harness exposes one bbox or nine.
The macro-average $C-V$ swings from $+0.10$ in the grid-marker family to $-0.06$ in the single-marker family---a 0.16 shift driven by the harness, not by the information condition.
Per-model behaviour is consistent: 9 of 12 models prefer C over V in the grid-marker family but 9 of 12 prefer V over C in the single-marker family.
This is the strongest within-task example in our benchmark of the broader pattern documented in Sec.~\ref{sec:case_studies}: the harness is a first-class confound, and a symbolic channel only helps when the interface presents the agent with multiple candidates that the channel can disambiguate.

\section{Maze Studies}
\label{app:maze_studies}

\subsection{Per-Model Breakdown}
\label{app:maze_results_section}

\begin{table}[h]
\centering
\caption{\textbf{MazeNavFPS results by information condition and action protocol.}
Pass@1 ($\uparrow$) for visual-only input (V) and rendered images with a structured position log (C: VP).
Batch (Bf) and single-action (Sa) control are reported separately.
Each mean uses all valid independent (grid, seed) configurations within that model and protocol; $n$ gives the actual count.
Repeated configurations contribute only their earliest valid attempt.
All runs use three wall materials and a budget of 50 model calls. Batch can execute several primitive actions per call, so the total primitive-action allowance is not matched to Single.}
\label{app:maze_results}
\small
\setlength{\tabcolsep}{4pt}
\begin{tabular}{l c c c c}
\toprule
\textbf{Model} & \textbf{V: Batch} & \textbf{V: Single} & \textbf{C: VP Batch} & \textbf{C: VP Single} \\
\midrule
Gemini-3.1-Pro Preview & 0.67 ($n=15$) & 0.11 ($n=9$) & 0.67 ($n=15$) & 0.11 ($n=9$) \\
Claude Sonnet-4.6 & 0.87 ($n=15$) & 0.44 ($n=9$) & 0.87 ($n=15$) & 0.22 ($n=9$) \\
GPT-5.2 & 0.46 ($n=13$) & 0.00 ($n=9$) & 0.79 ($n=14$) & 0.33 ($n=9$) \\
GPT-4.1 & 0.33 ($n=15$) & 0.00 ($n=7$) & 0.38 ($n=13$) & 0.22 ($n=9$) \\
Gemini-3-Flash Preview & 0.57 ($n=14$) & 0.13 ($n=8$) & 0.57 ($n=14$) & 0.00 ($n=9$) \\
Grok-4.3 & 0.67 ($n=9$) & 0.17 ($n=6$) & 0.43 ($n=14$) & 0.38 ($n=8$) \\
\bottomrule
\end{tabular}
\end{table}

\subsection{Material Texture Ablation on the 4$\times$4 Maze}
\label{app:material}

Table~\ref{tab:material} reports pass rates broken down by wall material on the 4$\times$4 maze (batch free, max 30 steps).
Three material configurations are tested: \textbf{mat1} (uniform single-color brick), \textbf{mat2} (two alternating materials), and \textbf{mat3} (three-material checkerboard pattern providing the strongest visual landmarks at corners).

\begin{table}[h]
  \centering
  \caption{
    \textbf{Material texture ablation} (4$\times$4 maze, batch free).
    Each cell shows pass rate, with average rounds in parentheses.
    Richer visual textures (mat3) tend to improve VP pass rates by providing distinguishable landmarks.
  }
  \label{tab:material}
  \small
  \setlength{\tabcolsep}{3pt}
  \begin{tabular}{@{}l cc cc cc@{}}
    \toprule
    & \multicolumn{2}{c}{\textbf{mat1 (uniform)}} & \multicolumn{2}{c}{\textbf{mat2 (2-tile)}} & \multicolumn{2}{c}{\textbf{mat3 (3-tile)}} \\
    \cmidrule(lr){2-3} \cmidrule(lr){4-5} \cmidrule(lr){6-7}
    \textbf{Model} & VP & VRP & VP & VRP & VP & VRP \\
    \midrule
    Gemini 3 Pro  & 1.00\,(9.00)  & 1.00\,(16.00) & 0.50\,(19.50) & 1.00\,(13.50) & 1.00\,(7.20)  & 1.00\,(8.00) \\
    Claude Opus   & 0.50\,(17.50) & 0.50\,(23.50) & 1.00\,(14.00) & 1.00\,(10.00) & 0.60\,(26.80) & 0.50\,(19.50) \\
    GPT-4.1       & 0.00\,(30.00) & 0.00\,(30.00) & 0.50\,(30.00) & 0.00\,(30.00) & 0.00\,(30.00) & 0.50\,(19.50) \\
    \bottomrule
  \end{tabular}
  \vspace{0.3em}

  {\footnotesize Format: pass rate (average rounds). Each cell has $n=2$, except mat3 VP: $n=4$ for Gemini and $n=5$ for Opus and GPT-4.1. VP = screenshot + position history; VRP = VP + raycast.}
\end{table}

Gemini achieves 1.00 on mat3 VP but drops to 0.50 on mat2 VP, suggesting the three-material checkerboard provides stronger visual cues for spatial orientation.
GPT-4.1 achieves a pass rate of 0.50 with VP on mat2 and with VRP on mat3.

\section{Failure Modes on Chaotic and Continuous-Dynamics Tasks}
\label{app:chaotic}

Tasks involving continuous physical dynamics---such as surface billiards, surface deformation, and trajectory prediction---expose a qualitatively different class of VLM failures compared to discrete navigation tasks. We analyze common patterns across these task families.

\paragraph{Sensitivity to initial conditions.}
In surface billiards, the ball trajectory on a curved Slang-defined surface is deterministic but chaotic: small differences in perceived initial position or heading lead to large differences in predicted landing position. No tested model predicts these trajectories reliably, even when the qualitative direction is approximately correct; the best CurvyBilliards score in Table~\ref{tab:core_results} is 0.28.

\paragraph{Visual-to-continuous-parameter mapping.}
Unlike discrete tasks where actions are categorical (turn left/right, fold $\pm 90^{\circ}$), continuous-dynamics tasks require mapping visual observations to real-valued parameters (surface slope, curvature coefficients, velocity vectors). This mapping demands quantitative spatial precision that current VLMs lack:
\begin{enumerate}
  \item \textbf{3D trajectory inference from 2D projection}: Models struggle to recover heading direction from a single camera viewpoint. Depth ambiguity causes systematic errors in distance and direction estimation.
  \item \textbf{Surface curvature reasoning}: Even when the heading is approximately correct, models fail to predict how parametric surface curvature deflects the trajectory---a computation requiring implicit integration of local geometry.
  \item \textbf{Limited refinement}: Unlike navigation tasks with step-by-step feedback, these tasks allow at most five submit-and-check rounds, so errors must be corrected from a few outcome observations.
\end{enumerate}

\paragraph{Implications for benchmark design.}
These failure modes suggest that chaotic and continuous-dynamics tasks represent a fundamentally harder regime for VLMs than discrete-action tasks. The visual-to-structure gap identified in the core tasks (where structured input narrows the gap) may not transfer straightforwardly: even with structured input, predicting chaotic dynamics requires numerical simulation capabilities that current VLMs do not possess. This motivates the code-as-action-space design in tasks like DropToTarget, where the agent writes a surface-height function and the harness evaluates it using GPU simulation.

\section{Running the Released Benchmark}
\label{app:implementation}

\subsection{Install the Binary and Python Tasks}

The benchmark runs from a packaged Unreal executable and the accompanying Python task code. Extract the complete Windows bundle to a writable directory, preserving its folder structure. Use an NVIDIA GPU with graphics support, a current driver, and Python~3.13. Running the supplied binary does not require Unreal Editor or rebuilding the engine. In PowerShell, from the extracted directory:
\begin{verbatim}
py -3.13 -m venv .venv
.\.venv\Scripts\python.exe -m pip install -r requirements.txt
Copy-Item model_configs.example.json model_configs.json
\end{verbatim}
The launcher automatically locates the binary inside the Windows bundle. For a separately downloaded runtime, set \texttt{VERIWORLD\_BUILD} or pass \texttt{-{}-build} with the executable path. Linux uses a separate runtime build and the same task-launcher arguments; platform setup is documented in the package README. MovingShooter, DropToTarget, and CurvyBilliards additionally require \texttt{ffmpeg} and \texttt{ffprobe} on the host's \texttt{PATH}.

\paragraph{Tested hardware.} Windows runs used NVIDIA GeForce RTX 3090 workstations, each running several Unreal instances in parallel (\texttt{run\_parallel.py}, default \texttt{max\_instances}=6). Linux runs used RunPod single-GPU instances, mostly NVIDIA A40 and also RTX 3090/4090/A5000/A6000, L4, L40/L40S and RTX PRO Blackwell GPUs, with one or two Unreal instances per GPU. A native smoke test with a scripted agent (no model API; MazeNavFPS level loading, rendering, movement and logging) passed with 1, 4 and 6 concurrent instances on each of NVIDIA A40, A100 (SXM4 80\,GB), H100 (80\,GB HBM3) and GeForce RTX 4090; with six instances, peak GPU memory was 21.7--23.4\,GB (about 4\,GB per instance).

\subsection{Configure the Model API}

Edit \texttt{model\_configs.json}: \texttt{name} is the local configuration label, \texttt{model} is the provider's requested model identifier, and \texttt{base\_url} is its API endpoint. Set \texttt{client\_options.expected\_model} to the expected returned model identifier. Keep \texttt{api\_key} as \texttt{env:VERIWORLD\_API\_KEY} and set that environment variable to your own credential in the shell used to launch the task. The example below uses the configuration label \texttt{example-model}. Provider-specific protocol and request options are documented in \texttt{API\_CONFIGURATION.txt}.

\subsection{Run a Task and Inspect the Results}

First check local dependencies and binary discovery, then run a single model--environment round:
{\small
\begin{verbatim}
$py = ".\.venv\Scripts\python.exe"
& $py run_task.py --check
& $py run_task.py --task PlaceOnBottom --seed 0 --model example-model --smoke
\end{verbatim}
}
The launcher starts the engine and connects the selected task to the model API. The \texttt{-{}-check} command makes no API request; \texttt{-{}-smoke} makes real model requests for a one-round installation check. Remove \texttt{-{}-smoke} for the task's normal budget, or specify \texttt{-{}-rounds}. Select tasks with \texttt{-{}-list}, choose instances with \texttt{-{}-seed}, and use \texttt{-{}-condition V}, \texttt{S}, or \texttt{C} where supported (\texttt{V-pure} runs the pixel-only controls for PlaceOnBottom, BoxFold and LightWar). To reproduce a reported comparison, use its recorded model, seed, information condition, harness, and budget; other harness variants are listed in \texttt{task\_entrypoints.json}.

Each run creates an output directory recording the invocation, model inputs, responses from successful API calls, rendered observations, actions, and verifier outcomes. Installation checks are kept separate from benchmark scores. API or runtime errors remain unscored; a valid action that fails the task verifier is a model failure. For concurrent runs, assign a distinct \texttt{-{}-port} and \texttt{-{}-output} to each instance. The package README provides the complete commands and output layout.

\paragraph{Artifact licensing and permitted use.}
The benchmark code and task definitions are publicly available under the licenses specified in the repository. The packaged VeriWorld runtime and bundled scene assets are provided solely for model evaluation and benchmark reproduction in research. This release does not grant permission to extract or repurpose the bundled assets or proprietary runtime components for game development or other standalone applications. Unreal Engine and third-party assets remain subject to their respective licenses, including the Unreal Engine EULA and applicable content licenses. No additional rights to these third-party components are granted by the benchmark release.

\section{Open-Source Platform and Verification}
\label{app:opensource}

Task metadata and seed fixtures are available on Hugging Face (\url{https://huggingface.co/datasets/yanyanzheng96/VeriWorld}) and GitHub (\repourl). These tasks require the packaged VeriWorld binary; the repository README gives its download location. The platform follows a three-layer design where each layer has a clear responsibility, and new tasks are authored by filling in task-specific logic at the appropriate layer without modifying infrastructure.

\subsection{Platform Architecture}

\paragraph{Layer 1: Skills + Framework (\texttt{veriworld/infra/}).}
Everything a task author needs to get started. Two engine classes serve different task categories:
\begin{itemize}[leftmargin=*, itemsep=1pt]
  \item \texttt{InteractiveEngine}: Persistent UE5 instance with WebSocket connection for per-tick control (screenshot $\to$ think $\to$ act $\to$ screenshot). Handles process launch/teardown, level loading, and exposes \texttt{python\_exec()}, \texttt{screenshot()}, and raw JSON-RPC via the underlying \texttt{UEClient}.
  \item \texttt{ComputationalEngine}: Per-round UE restart for GPU shader tasks (Slang/CUDA compute).
\end{itemize}
Both engines are paired with annotated \texttt{task\_template.py} skeletons containing inline documentation and a complete worked example (MazeNavFPS for interactive, SurfaceBilliards for computational) in comments. VeriWorld-authored skill documents cover Slang compute shader conventions (\texttt{infra/slang/SKILL.md}) and Lean formal verification harness (\texttt{infra/lean/SKILL.md}).

\paragraph{Layer 2: Task Parameterization + Solvability (\texttt{veriworld/benchmark/*/}).}
A typical task package (e.g., MazeNavFPS) contains:
\begin{itemize}[leftmargin=*, itemsep=1pt]
  \item \texttt{generate\_params.py}: Pure, seeded scene generator. All scene-level ablation axes (grid size, wall materials, tunnel radius, start/goal yaw) are parameters here. Seed determinism guarantees reproducibility across runs and models.
  \item \texttt{ue\_setup.py} + \texttt{move\_camera.py}: UE-side scene builder and per-tick movement loop, shared across all ablations of the task.
  \item Where applicable, a \emph{solvability check}: either (a) a closed-form check (maze goal distance), (b) a Slang compute shader that simulates ground-truth physics deterministically (\texttt{billiard\_ball.slang}), or (c) a Lean~4 per-seed reachability theorem (\texttt{native\_decide}) confirming the instantiated parameters admit a solution. Without such a guarantee, ablation results are meaningless---``the model failed'' is not a claim unless the instance is known to be solvable.
\end{itemize}

\paragraph{Layer 3: Agent-Harness Ablations (per-task subfolders).}
Sibling subfolders within each task encode fundamentally different experimental conditions, each usually a self-contained \texttt{task.py}:
\begin{itemize}[leftmargin=*, itemsep=1pt]
  \item \texttt{mazenavfps/harness\_structured/vp\_bf/}: Screenshot + position log, batch-free commands.
  \item \texttt{mazenavfps/harness\_structured/pv\_bf/}: Pure vision history grid, no coordinates, BLOCKED/ok only.
  \item \texttt{tunnel\_nav3d/harness\_structured/af/}: Compound aim-and-fly action space.
\end{itemize}
A new subfolder is warranted when the ablation changes action space, protocol (per-tick/batch/per-round), or knowledge organization. Changes that only vary scene parameters belong in Layer~2 as CLI flags.

\paragraph{Fair-comparison parallelism.}
The orchestrator (\texttt{veriworld/scripts/run\_parallel.py}) enforces that each seed runs on every participating model simultaneously across parallel UE instances. \texttt{max\_instances} (default 6, matching single-GPU VRAM budget) caps concurrency; batches are rounded to multiples of $N_\text{models}$ so no seed straddles batches. Model participation is controlled by \texttt{parallel: true/false} in \texttt{model\_configs.json}, overridable with \texttt{-{}-models}.

\subsection{Verification Examples: Lean, Slang, and Python}
\label{app:verify_examples}

\ours uses three verification technologies depending on task complexity, described below.

\subsubsection{Lean 4: Maze Connectivity Proof}

For the maze navigation task, the \lean proof shows that a path exists from start to goal. It mirrors the Python DFS generator with an LCG-based reference RNG and proves connectivity via BFS reachability.

For each specified seed, \texttt{native\_decide} checks the generated grid\textquotesingle s Boolean connectivity predicate. The \href{https://github.com/VITA-Group/VeriWorld/blob/paper-artifacts-2026-11-06-rc3/veriworld/benchmark/interactive_control/mazenavfps/lean_verify/MazeConnectivity.lean}{released Lean source} contains the grid definitions and per-seed theorems.

The released theorems cover seeds 0, 1, 2, 3, 5, 7, 10, 42 and 99 on $3{\times}3$ and $4{\times}4$ grids and seeds 0, 7 and 42 on $5{\times}5$ grids. Mazes for other seeds are connected by construction, because depth-first carving visits every cell. A companion script (\texttt{cross\_check.py}) checks that the Python generator reproduces the Lean reference grids for pinned seeds.

\subsubsection{Slang Compute Shader: Ball Physics on Parametric Surface}

The DropToTarget and SurfaceBilliards tasks use Slang GPU shaders for deterministic physics verification. The ball finds its nearest surface node in 3D, computes the surface normal from neighbours, and projects gravity onto the tangent plane.

The DropToTarget kernel uses \texttt{dt}$=0.003$, \texttt{mu}$=0.4$, and \texttt{grav}$=300.0$; the CurvyBilliards kernels use lower friction. The \href{https://github.com/VITA-Group/VeriWorld/tree/paper-artifacts-2026-11-06-rc3/veriworld/benchmark/computational_physics}{released GPU implementations} provide the task-specific simulation kernels.

The harness runs the Slang shader deterministically on the GPU to evaluate the submitted action. In DropToTarget, the agent writes a surface-height function, and the harness simulates the resulting ball motion and checks whether the ball's first ground contact falls within the target circle.

\subsubsection{Python: BFS Ground Truth Oracle}

For tasks where the verification criterion is straightforward (e.g., ``did the agent reach the goal cell?''), a pure Python oracle suffices.

The oracle explores traversable cells in breadth-first order, returning a start-to-goal path when reachable and \texttt{None} otherwise. See the \href{https://github.com/VITA-Group/VeriWorld/blob/paper-artifacts-2026-11-06-rc3/veriworld/benchmark/interactive_control/mazenavfps/lean_verify/ground_truth.py}{released Python oracle} for the implementation.

These checks (Lean per-seed reachability theorems, Slang physics simulation, Python closed-form oracles) certify that a solution exists for the instances they cover; as described in Sec.~\ref{sec:230}, task families whose construction guarantees a solution are not separately certified, and no check certifies completion within the step budget.

\clearpage
\section{Supplementary Experiments and Analyses}
\label{app:supplementary_experiments}

For the PlaceOnBottom and BoxFold rows and Gemini-3.1-Pro on MovingShooter in Table~\ref{tab:core_results}, each condition uses all available valid seeds. Table~\ref{tab:supp_main_counts} gives the per-condition sample counts; the matched analyses below use the intersection of valid seeds within each model.
\par\medskip\noindent\begin{minipage}{\linewidth}\centering\small
\captionof{table}{Sample counts for the six main-table rows averaged over all valid seeds per condition.}\label{tab:supp_main_counts}
\setlength{\tabcolsep}{5pt}
\begin{tabular}{@{}llrrr@{}}\toprule
Task & Model & $n_V$ & $n_S$ & $n_C$ \\ \midrule
PlaceOnBottom & Gemini-3.1-Pro & 20 & 50 & 50 \\
PlaceOnBottom & Opus-4.5 & 49 & 21 & 21 \\
BoxFold & Gemini-3.1-Pro & 20 & 30 & 21 \\
BoxFold & Flash-2.5 & 29 & 21 & 21 \\
BoxFold & Sonnet-4.6 & 42 & 20 & 21 \\
MovingShooter & Gemini-3.1-Pro & 20 & 40 & 29 \\
\bottomrule\end{tabular}\end{minipage}\par\medskip

\paragraph{Billiards rows.} Table~\ref{tab:core_results} lists three representative \textit{Billiards} models; Table~\ref{tab:supp_billiards} gives all nine.
\par\medskip\noindent\begin{minipage}{\linewidth}\centering\small
\captionof{table}{\textit{Billiards}: all nine models in Table~\ref{tab:core_results} (standard preset, mean fraction of balls pocketed, 3 seeds).}\label{tab:supp_billiards}
\setlength{\tabcolsep}{5pt}
\begin{tabular}{@{}lrrrrr@{}}\toprule
Model & V & S & C & S$-$V & C$-\max$(V,S) \\ \midrule
Gemini-3-Pro & 0.33 & 0.67 & 0.53 & +0.33 & -0.13 \\
Flash        & 0.33 & 0.60 & 0.73 & +0.27 & +0.13 \\
Sonnet       & 0.20 & 0.47 & 0.53 & +0.27 & +0.07 \\
GPT-5.2      & 0.13 & 0.40 & 0.27 & +0.27 & -0.13 \\
GPT-4.1      & 0.13 & 0.27 & 0.53 & +0.13 & +0.27 \\
Qwen3-235B   & 0.00 & 0.33 & 0.13 & +0.33 & -0.20 \\
Doubao       & 0.13 & 0.27 & 0.40 & +0.13 & +0.13 \\
Moonshot     & 0.20 & 0.20 & 0.13 &  0.00 & -0.07 \\
Grok         & 0.27 & 0.33 & 0.33 & +0.07 &  0.00 \\
\bottomrule\end{tabular}\end{minipage}\par\medskip

\subsection{Matched information-condition panels}
\label{app:supp_matched_panels}
A seed specifies a task initialization, an information condition specifies the input available to the model, and an episode is one scored interaction under a fixed harness. Tables~\ref{tab:supp_PlaceOnBottom}--\ref{tab:supp_LightWar} report supplementary panels with matched V/S/C seeds within each model. These matched-seed panels are reported separately from the per-condition averages in Table~\ref{tab:core_results}. The larger LightWar comparison in Appendix~\ref{app:harness_lightwar} is also a separate panel.

We deduplicate copied records and exclude episodes with an explicit API or runtime failure or an incomplete response sequence. Each seed contributes one valid episode per condition; other valid repeated runs remain in the record archive. Each three-condition comparison uses the intersection of valid seeds, while valid outcomes without all three partners are retained separately. Missing outcomes are not scored as model failures. The accompanying record manifest preserves the selected source hashes and API identifiers. Generated parameters and interaction budgets are matched across V/S/C within each model. Macro-averages give equal weight to each model. Wilson intervals use confidence level 0.95 and describe binary-outcome sampling uncertainty within the listed configuration.
\par\medskip\noindent\begin{minipage}{\linewidth}\centering\small
\captionof{table}{Supplementary paired-panel sizes. The final column counts all three input conditions; seeds per condition are matched separately within each model.}\label{tab:supp_counts}
\setlength{\tabcolsep}{4pt}
\begin{tabular}{@{}lrrrr@{}}\toprule
Task & Models & Seeds per condition & Round budget & Episodes \\ \midrule
PlaceOnBottom & 11 & 16--21 & 8 & 633 \\
BoxFold & 12 & 7--21 & 15 & 666 \\
MovingShooter & 10 & 16--20 & 5 & 588 \\
LightWar & 10 & 7--10 & 8 & 285 \\
\bottomrule\end{tabular}\end{minipage}\par\medskip

\par\medskip\noindent\begin{minipage}{\linewidth}\centering\footnotesize
\captionof{table}{PlaceOnBottom: success rates and Wilson intervals. Each model uses the same $n$ seeds in V, S, and C.}\label{tab:supp_PlaceOnBottom}
\setlength{\tabcolsep}{4pt}
\begin{tabular}{@{}lrccc@{}}\toprule
Model & $n$ & V & S & C \\ \midrule
Claude Opus 4.5 & 21 & 0.33 [0.17, 0.55] & 1.00 [0.85, 1.00] & 1.00 [0.85, 1.00] \\
Claude Sonnet 4.6 & 20 & 0.35 [0.18, 0.57] & 1.00 [0.84, 1.00] & 1.00 [0.84, 1.00] \\
Doubao Seed 1.6 Flash & 18 & 0.56 [0.34, 0.75] & 0.83 [0.61, 0.94] & 0.89 [0.67, 0.97] \\
Gemini 2.5 Flash & 20 & 0.30 [0.15, 0.52] & 1.00 [0.84, 1.00] & 1.00 [0.84, 1.00] \\
Gemini 3.1 Pro Preview & 20 & 0.85 [0.64, 0.95] & 1.00 [0.84, 1.00] & 1.00 [0.84, 1.00] \\
GPT-4o & 20 & 0.55 [0.34, 0.74] & 1.00 [0.84, 1.00] & 1.00 [0.84, 1.00] \\
GPT-5-mini & 17 & 0.53 [0.31, 0.74] & 1.00 [0.82, 1.00] & 1.00 [0.82, 1.00] \\
GPT-5.2 & 20 & 0.20 [0.08, 0.42] & 1.00 [0.84, 1.00] & 1.00 [0.84, 1.00] \\
Kimi K2.5 & 19 & 0.37 [0.19, 0.59] & 1.00 [0.83, 1.00] & 1.00 [0.83, 1.00] \\
Qwen3-Max & 16 & 0.44 [0.23, 0.67] & 0.94 [0.72, 0.99] & 1.00 [0.81, 1.00] \\
Qwen3-VL-30B-A3B & 20 & 0.15 [0.05, 0.36] & 0.55 [0.34, 0.74] & 0.25 [0.11, 0.47] \\
\midrule
Macro-average & -- & 0.42 & 0.94 & 0.92 \\
\bottomrule\end{tabular}\end{minipage}\par\medskip

\par\medskip\noindent\begin{minipage}{\linewidth}\centering\footnotesize
\captionof{table}{BoxFold: success rates and Wilson intervals. Each model uses the same $n$ seeds in V, S, and C.}\label{tab:supp_BoxFold}
\setlength{\tabcolsep}{4pt}
\begin{tabular}{@{}lrccc@{}}\toprule
Model & $n$ & V & S & C \\ \midrule
Claude Opus 4.5 & 19 & 0.11 [0.03, 0.31] & 0.74 [0.51, 0.88] & 0.63 [0.41, 0.81] \\
Claude Sonnet 4.6 & 20 & 0.00 [0.00, 0.16] & 0.40 [0.22, 0.61] & 0.35 [0.18, 0.57] \\
Doubao Seed 1.6 Flash & 20 & 0.00 [0.00, 0.16] & 0.15 [0.05, 0.36] & 0.10 [0.03, 0.30] \\
Gemini 2.5 Flash & 21 & 0.00 [0.00, 0.15] & 0.33 [0.17, 0.55] & 0.33 [0.17, 0.55] \\
Gemini 3.1 Pro Preview & 20 & 0.60 [0.39, 0.78] & 1.00 [0.84, 1.00] & 1.00 [0.84, 1.00] \\
GPT-4o & 20 & 0.00 [0.00, 0.16] & 0.15 [0.05, 0.36] & 0.00 [0.00, 0.16] \\
GPT-5-mini & 17 & 0.00 [0.00, 0.18] & 0.35 [0.17, 0.59] & 0.29 [0.13, 0.53] \\
GPT-5.2 & 19 & 0.00 [0.00, 0.17] & 0.95 [0.75, 0.99] & 0.32 [0.15, 0.54] \\
Grok 4.3 & 20 & 0.00 [0.00, 0.16] & 0.70 [0.48, 0.85] & 0.45 [0.26, 0.66] \\
Kimi K2.5 & 20 & 0.00 [0.00, 0.16] & 0.45 [0.26, 0.66] & 0.05 [0.01, 0.24] \\
Qwen3-Max & 7 & 0.00 [0.00, 0.35] & 0.57 [0.25, 0.84] & 0.00 [0.00, 0.35] \\
Qwen3-VL-30B-A3B & 19 & 0.00 [0.00, 0.17] & 0.00 [0.00, 0.17] & 0.00 [0.00, 0.17] \\
\midrule
Macro-average & -- & 0.06 & 0.48 & 0.29 \\
\bottomrule\end{tabular}\end{minipage}\par\medskip

\par\medskip\noindent\begin{minipage}{\linewidth}\centering\footnotesize
\captionof{table}{MovingShooter: success rates and Wilson intervals. Each model uses the same $n$ seeds in V, S, and C.}\label{tab:supp_MovingShooter}
\setlength{\tabcolsep}{4pt}
\begin{tabular}{@{}lrccc@{}}\toprule
Model & $n$ & V & S & C \\ \midrule
Claude Sonnet 4.6 & 20 & 0.00 [0.00, 0.16] & 0.30 [0.15, 0.52] & 0.10 [0.03, 0.30] \\
Doubao Seed 1.6 Flash & 16 & 0.00 [0.00, 0.19] & 0.00 [0.00, 0.19] & 0.00 [0.00, 0.19] \\
Gemini 2.5 Flash & 20 & 0.00 [0.00, 0.16] & 0.00 [0.00, 0.16] & 0.05 [0.01, 0.24] \\
Gemini 3.1 Pro Preview & 20 & 0.00 [0.00, 0.16] & 0.35 [0.18, 0.57] & 0.10 [0.03, 0.30] \\
GPT-4o & 20 & 0.00 [0.00, 0.16] & 0.00 [0.00, 0.16] & 0.00 [0.00, 0.16] \\
GPT-5-mini & 20 & 0.00 [0.00, 0.16] & 0.00 [0.00, 0.16] & 0.00 [0.00, 0.16] \\
GPT-5.2 & 20 & 0.00 [0.00, 0.16] & 0.45 [0.26, 0.66] & 0.45 [0.26, 0.66] \\
Kimi K2.5 & 20 & 0.00 [0.00, 0.16] & 0.00 [0.00, 0.16] & 0.05 [0.01, 0.24] \\
Qwen3-Max & 20 & 0.00 [0.00, 0.16] & 0.05 [0.01, 0.24] & 0.05 [0.01, 0.24] \\
Qwen3-VL-30B-A3B & 20 & 0.00 [0.00, 0.16] & 0.00 [0.00, 0.16] & 0.00 [0.00, 0.16] \\
\midrule
Macro-average & -- & 0.00 & 0.12 & 0.08 \\
\bottomrule\end{tabular}\end{minipage}\par\medskip

\par\medskip\noindent\begin{minipage}{\linewidth}\centering\footnotesize
\captionof{table}{LightWar: success rates and Wilson intervals. Each model uses the same $n$ seeds in V, S, and C.}\label{tab:supp_LightWar}
\setlength{\tabcolsep}{4pt}
\begin{tabular}{@{}lrccc@{}}\toprule
Model & $n$ & V & S & C \\ \midrule
Claude Sonnet 4.6 & 10 & 0.10 [0.02, 0.40] & 1.00 [0.72, 1.00] & 1.00 [0.72, 1.00] \\
Doubao Seed 1.6 Flash & 9 & 0.00 [0.00, 0.30] & 1.00 [0.70, 1.00] & 1.00 [0.70, 1.00] \\
Gemini 2.5 Flash & 10 & 0.00 [0.00, 0.28] & 0.90 [0.60, 0.98] & 0.80 [0.49, 0.94] \\
Gemini 3.1 Pro Preview & 10 & 0.80 [0.49, 0.94] & 1.00 [0.72, 1.00] & 1.00 [0.72, 1.00] \\
GPT-4o & 10 & 0.00 [0.00, 0.28] & 0.50 [0.24, 0.76] & 0.00 [0.00, 0.28] \\
GPT-5-mini & 10 & 0.00 [0.00, 0.28] & 1.00 [0.72, 1.00] & 1.00 [0.72, 1.00] \\
GPT-5.2 & 9 & 0.00 [0.00, 0.30] & 1.00 [0.70, 1.00] & 1.00 [0.70, 1.00] \\
Kimi K2.5 & 10 & 0.00 [0.00, 0.28] & 1.00 [0.72, 1.00] & 0.90 [0.60, 0.98] \\
Qwen3-Max & 10 & 0.00 [0.00, 0.28] & 1.00 [0.72, 1.00] & 1.00 [0.72, 1.00] \\
Qwen3-VL-30B-A3B & 7 & 0.00 [0.00, 0.35] & 0.71 [0.36, 0.92] & 1.00 [0.65, 1.00] \\
\midrule
Macro-average & -- & 0.09 & 0.91 & 0.87 \\
\bottomrule\end{tabular}\end{minipage}\par\medskip

\subsection{Information condition crossed with action span}
\label{app:supp_span}
We cross V/S input with single-command and batched-command configurations in PlaceOnBottom. The single-command setting allows eight rounds. The batch setting allows three rounds, at most five commands in one batch, and six primitive actions in total. V and S share the same budget within each setting. Table~\ref{tab:supp_span} uses seeds available in all four arms. All six models retain a positive S$-$V gap under both settings. Because the two settings have different total action budgets, the between-setting difference describes the tested configurations rather than an isolated effect of batching alone.
\par\medskip\noindent\begin{minipage}{\linewidth}\centering\footnotesize
\captionof{table}{PlaceOnBottom success rates under two action spans. Both information-condition contrasts are positive for every listed model.}\label{tab:supp_span}
\setlength{\tabcolsep}{4pt}
\begin{tabular}{@{}lrrrrrrr@{}}\toprule
Model & $n$ & Single V & Single S & Batch V & Batch S & $\Delta_{\rm single}$ & $\Delta_{\rm batch}$ \\ \midrule
Claude Sonnet 4.6 & 20 & 0.35 & 1.00 & 0.35 & 1.00 & +0.65 & +0.65 \\
Gemini 2.5 Flash & 20 & 0.30 & 1.00 & 0.20 & 1.00 & +0.70 & +0.80 \\
Gemini 3.1 Pro Preview & 20 & 0.85 & 1.00 & 0.35 & 1.00 & +0.15 & +0.65 \\
GPT-5-mini & 17 & 0.53 & 1.00 & 0.35 & 1.00 & +0.47 & +0.65 \\
GPT-5.2 & 20 & 0.20 & 1.00 & 0.35 & 1.00 & +0.80 & +0.65 \\
Qwen3-VL-30B-A3B & 20 & 0.15 & 0.55 & 0.00 & 0.10 & +0.40 & +0.10 \\
\bottomrule\end{tabular}\end{minipage}\par\medskip

\subsection{Prompt-length and serialization controls}
\label{app:supp_prompt_controls}
For the prompt-length control, we append 750 words to the structured input. One variant uses repeated filler; the other uses coherent task-irrelevant prose. Both padded variants allow six rounds, while the S and C baselines allow eight. Table~\ref{tab:supp_padding} matches seeds across all four arms. Natural-language padding changes the number of successful episodes by at most one per model. The repetitive filler produces a larger change for Sonnet, from 1.00 to 0.35. The results distinguish the tested filler constructions; they do not support treating all prompt extensions as equivalent.
\par\medskip\noindent\begin{minipage}{\linewidth}\centering\small
\captionof{table}{PlaceOnBottom padding controls. Entries are success rates on the common valid seeds.}\label{tab:supp_padding}
\setlength{\tabcolsep}{4pt}
\begin{tabular}{@{}lrrrrr@{}}\toprule
Model & $n$ & S & Repeated filler & Natural prose & C \\ \midrule
Claude Sonnet 4.6 & 20 & 1.00 & 0.35 & 1.00 & 1.00 \\
Doubao Seed 1.6 Flash & 19 & 0.84 & 0.84 & 0.79 & 0.84 \\
Gemini 2.5 Flash & 20 & 1.00 & 1.00 & 1.00 & 1.00 \\
Gemini 3.1 Pro Preview & 20 & 1.00 & 1.00 & 1.00 & 1.00 \\
GPT-4o & 20 & 1.00 & 1.00 & 1.00 & 1.00 \\
GPT-5-mini & 20 & 1.00 & 1.00 & 1.00 & 1.00 \\
GPT-5.2 & 20 & 1.00 & 1.00 & 1.00 & 1.00 \\
Kimi K2.5 & 20 & 1.00 & 1.00 & 1.00 & 1.00 \\
Qwen3-Max & 19 & 0.95 & 1.00 & 0.95 & 1.00 \\
Qwen3-VL-30B-A3B & 20 & 0.55 & 0.65 & 0.55 & 0.25 \\
\bottomrule\end{tabular}\end{minipage}\par\medskip

A separate serialization control changes the order of objects and keys in the structured prompt while retaining its values and the eight-round budget. Three models change by at most one success, while Qwen3-VL-30B-A3B gains six successes on twenty matched seeds (Table~\ref{tab:supp_serialization}). Prompt organization can therefore affect performance even when the supplied scene state is unchanged.
\par\medskip\noindent\begin{minipage}{\linewidth}\centering\small
\captionof{table}{PlaceOnBottom serialization control with the same task state and round budget.}\label{tab:supp_serialization}
\setlength{\tabcolsep}{4pt}
\begin{tabular}{@{}lrrr@{}}\toprule
Model & $n$ & S & Reordered S \\ \midrule
Claude Sonnet 4.6 & 20 & 1.00 & 1.00 \\
Doubao Seed 1.6 Flash & 19 & 0.84 & 0.89 \\
Qwen3-Max & 20 & 0.95 & 1.00 \\
Qwen3-VL-30B-A3B & 20 & 0.55 & 0.85 \\
\bottomrule\end{tabular}\end{minipage}\par\medskip

\subsection{Combined input: an episode example and instruction controls}
\label{app:supp_combined_example}
LightWar supplies launch and target coordinates in both S and C. First-round actions can be compared before different shots create different interaction histories. In one GPT-4o episode, the C response writes an arctangent using coordinate differences 36,122 and 49,962 but reports a bearing of $59.90^\circ$. The calculation gives $35.87^\circ$; under S, the model reports $35.90^\circ$ on the same instance. This is a concrete example of an incorrect calculation despite the relevant state being present.

The original C instruction asks the model to estimate yaw visually and cross-check the coordinates. We compare it with a neutral instruction that directs the model to the structured state, and an instruction to ignore vision while retaining the images. Table~\ref{tab:supp_lightwar_instruction} reports ten matched seeds per condition. The neutral instruction improves success from 0.00 to 0.60 without removing the image. This pattern shows that the prompt's direction of attention to the two channels contributes to the observed behavior; the episode alone does not identify a general mechanism of multimodal fusion.
\par\medskip\noindent\begin{minipage}{\linewidth}\centering\footnotesize
\captionof{table}{LightWar instruction controls for GPT-4o. Bearing error uses the first-round action. One neutral-condition reply has no valid first-round action; that episode remains in the success-rate denominator, and its bearing error is undefined.}\label{tab:supp_lightwar_instruction}
\setlength{\tabcolsep}{4pt}
\begin{tabular}{@{}lrrrr@{}}\toprule
Condition & $n$ & Success & Bearings & Median error \\ \midrule
S: structured input & 10 & 0.50 & 10 & $0.83^\circ$ \\
C: visual estimate, then cross-check & 10 & 0.00 & 10 & $142.59^\circ$ \\
C: neutral structured-state instruction & 10 & 0.60 & 9 & $1.40^\circ$ \\
C: ignore-vision instruction & 10 & 0.30 & 10 & $4.37^\circ$ \\
\bottomrule\end{tabular}\end{minipage}\par\medskip

\subsection{Harness ranking stability}
\label{app:supp_rank}
Kendall's $\tau_b$ measures agreement between two model rankings while accounting for ties. Table~\ref{tab:supp_rank} recomputes this statistic from the reported scores in the existing harness tables, using the same model set within each pair. The Billiards comparison uses mean fraction pocketed; the others use episode success rates. Across these five pairs, $\tau_b$ ranges from 0.61 to 0.83. Thus substantial changes in absolute scores can coexist with positive ranking agreement.
\par\medskip\noindent\begin{minipage}{\linewidth}\centering\small
\captionof{table}{Ranking agreement across the existing harness pairs. The ObjectGrounding rows use the twelve-model tables in Appendix~\ref{app:harness_object_ground}.}\label{tab:supp_rank}
\setlength{\tabcolsep}{4pt}
\begin{tabular}{@{}lrr@{}}\toprule
Comparison & Models & Kendall $\tau_b$ \\ \midrule
Billiards V/REPL & 11 & +0.76 \\
MazeNavFPS VP, 50 calls & 6 & +0.75 \\
MazeNavFPS VRP, 50 calls & 7 & +0.79 \\
ObjectGrounding V & 12 & +0.83 \\
ObjectGrounding C & 12 & +0.61 \\
\bottomrule\end{tabular}\end{minipage}\par\medskip

\end{document}